\documentclass[journal]{IEEEtran}
\ifCLASSOPTIONcompsoc
  \usepackage[nocompress]{cite}
\else
  \usepackage{cite}
\fi
\usepackage{amsmath}
\usepackage{amssymb}
\usepackage{algorithm}
\usepackage{algorithmic}
\usepackage{amsfonts}
\usepackage{enumerate}
\usepackage{epsfig}
\usepackage{graphicx}
\usepackage{balance}

\usepackage{color}
\usepackage{soul}

\def\ie{{\em i.e.}}
\def\eg{{\em e.g.}}
\def\etal{{\em et al.}}

\newcommand{\figref}[1]{Fig. \ref{#1}}
\newcommand{\tabref}[1]{Tab. \ref{#1}}
\newcommand{\equref}[1]{(\ref{#1})}
\newcommand{\secref}[1]{Section \ref{#1}}

\newcommand{\mc}[1]{\mathcal{#1}}

\newcommand{\br}[1]{\bm{\mathrm{#1}}}

\graphicspath{{./figure/}}

\usepackage{ragged2e}
\usepackage{booktabs}
\usepackage{color}
\usepackage{multirow}
\usepackage{bm}
\usepackage{bbm}

\begin{document}
\title{RGB-D Salient Object Detection with Ubiquitous Target Awareness}

\author{Yifan~Zhao$^{\dagger}$,~Jiawei~Zhao$^{\dagger}$,~Jia~Li,~\IEEEmembership{Senior Member,~IEEE}, and~Xiaowu~Chen,~\IEEEmembership{Senior Member,~IEEE}% <-this % stops a space
\IEEEcompsocitemizethanks{\IEEEcompsocthanksitem
$^{\dagger}$Y. Zhao and J. Zhao contribute equally to this work.
\IEEEcompsocthanksitem Y. Zhao, J. Zhao, J. Li and X. Chen are with the State Key Laboratory of Virtual Reality Technology and Systems, School of Computer Science and Engineering, Beihang University, Beijing, 100191, China.
\IEEEcompsocthanksitem J. Li and X. Chen are also with the Peng Cheng Laboratory, Shenzhen, 518000, China.
\IEEEcompsocthanksitem J. Li is the corresponding author (E-mail: jiali@buaa.edu.cn).
\IEEEcompsocthanksitem A preliminary version of this research appeared at ACM MM 2020~\cite{oursMM}.
}}

\markboth{}%
{Zhao \MakeLowercase{\textit{et al.}}: }

\maketitle

\begin{abstract}
 Conventional RGB-D salient object detection methods aim to leverage depth as complementary information to find the salient regions in both modalities. However, the salient object detection results heavily rely on the quality of captured depth data which sometimes are unavailable. In this work, we make the first attempt to solve the RGB-D salient object detection problem with a novel depth-awareness framework. This framework only relies on RGB data in the testing phase, utilizing captured depth data as supervision for representation learning. To construct our framework as well as achieving accurate salient detection results, we propose a Ubiquitous Target Awareness (UTA) network to solve three important challenges in RGB-D SOD task: 1) a depth awareness module to excavate depth information and to mine ambiguous regions via adaptive depth-error weights, 2) a spatial-aware cross-modal interaction and a channel-aware cross-level interaction, exploiting the low-level boundary cues and amplifying high-level salient channels, and 3) a gated multi-scale predictor module to perceive the object saliency in different contextual scales. Besides its high performance, our proposed UTA network is depth-free for inference and runs in real-time with 43 FPS. Experimental evidence demonstrates that our proposed network not only surpasses the state-of-the-art methods on five public RGB-D SOD benchmarks by a large margin, but also verifies its extensibility on five public RGB SOD benchmarks.
 	
\end{abstract}

\begin{IEEEkeywords}
RGB-D Salient Object Detection, Depth-awareness, Real-time, Ubiquitous Target Awareness.
\end{IEEEkeywords}

\IEEEpeerreviewmaketitle

\section{Introduction}

Salient object detection (SOD) aims at detecting and segmenting objects that attract human attention most visually. With the proposals of large datasets~\cite{ju2014depth,peng2014rgbd,niu2012leveraging,yan2013hierarchical,li2015visual,wang2017learning} and deep learning techniques~\cite{he2016deep,long2015fully}, recent works have made significant progress in accurately segmenting salient objects,
which can serve as an important prerequisite for a wide range of computer vision tasks, including semantic segmentation~\cite{lai2016saliency}, visual tracking~\cite{hong2015online}, and image retrieval~\cite{shao2006specific}.

Recent years have witnessed significant progress in the field of salient object detection. Previous works~\cite{liu2018picanet,li2019constrained,wu2019cascaded,li2021salient,zhao2019pyramid,su2019selectivity,qin2019basnet,F3Net,li2020parallel,ma2021pyramidal} take only the RGB information as inputs, which are relatively lightweight and can be trained end-to-end.
However, the reasoning of salient regions cannot be well solved when there exist multiple contrasting region proposals or ambiguous object contours. Therefore, the depth information can serve as complementary guidance to deduct the overlapping objects, which is beneficial to salient object detection tasks.
Combing the RGB information with auxiliary depth inputs, recent research efforts~\cite{qu2017rgbd,chen2018progressively,chen2019multi,chen2019three,
zhang2020select,fan2019D3Net,chen2020dpanet,li2020asif,zhang2020uc,li2020rgb,li2020icnet,chen2020rgbd,Wei2020ECCV,oursMM,bai2021circular,zhao2019contrast,zhang2020uc,fan2020bbs} have verified its effectiveness in improving the object segmentation process.
These methods usually introduce an additional depth stream to encode depth maps and then fuse the RGB stream with depth stream to deduct the salient objects. For example, Piao \etal~\cite{piao2019depth} propose a two-stream network and fuse paired multi-level features to refine the final saliency results. Zhao \etal~\cite{zhao2019pyramid} propose a fluid pyramid integration strategy to exploit enhanced depth features.
Although promising improvements have been made, existing works mainly focus on the extraction of depth information and fusion strategies of multiple modalities. Rethinking this problem from another perspective, we, in particular, consider three challenges in RGB-D salient object detection.

%For example, Wu \etal ~\cite{wu2019cascaded} propose a coarse-to-fine feature aggregation framework to generate saliency maps.

\begin{figure*}[t]
	\begin{center}
		%\fbox{\rule{0pt}{2in} \rule{.9\linewidth}{0pt}}	
		\includegraphics[width= \linewidth]{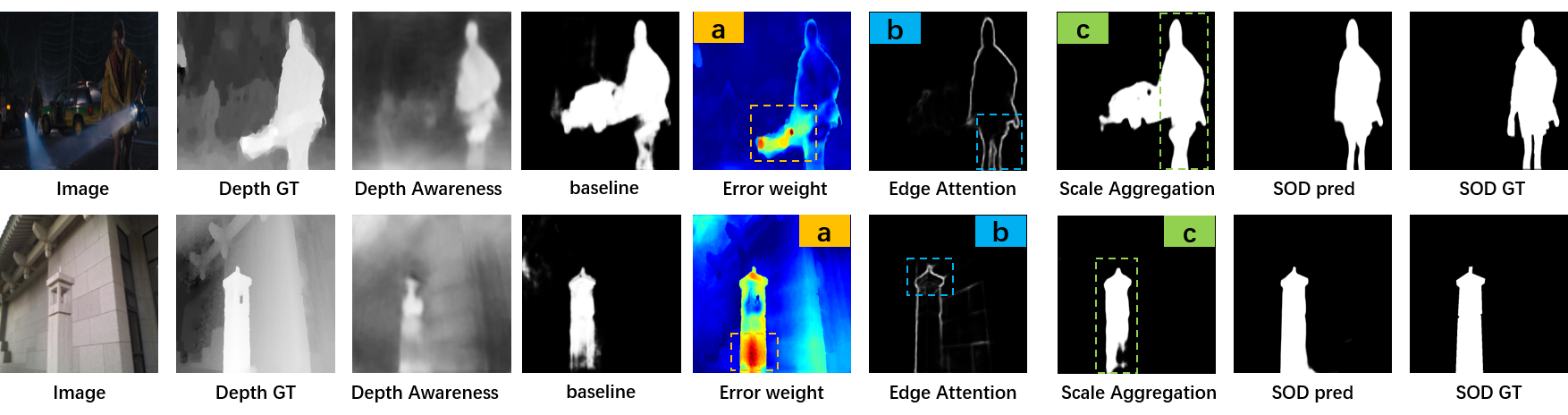}
		\caption{Illustration of our Ubiquitous Target Awareness (UTA) network, consisting of three meaningful designs. (a) depth awareness: focusing on depth regions and mining ambiguous regions by the generated error weights between predicted depth and depth groundtruth.
		(b) low-level cues awareness: embedding the edge information for cross-modal fusion.
		(c) scale awareness: perceiving multi-scale contextual information for saliency detection.
		}\label{fig:motivation}
	\end{center}
\end{figure*}

The first challenge is how to make the SOD network be aware of depth information. There exist some drawbacks in conventional RGB-D SOD methods. On the one hand, these conventional methods usually rely on auxiliary branches to extract depth features and then fuse these features with RGB ones. This would lead to heavy computation burdens and insufficient modality fusion.
On the other hand, object segmentation processes heavily rely on the acquisition of depth maps, which are unavailable on some extreme occasions or realistic applications.
Unlike these methods, we make the first attempt to leverage depth data as additional training constraints, which do not rely on depth inputs in the testing phase, namely depth-awareness salient object detection. Taking the essence and discarding the dregs of RGB and RGB-D methods, we propose a depth-awareness module to regularize network features being aware of depth knowledge. This can be conducted in a multi-task learning trend when learning saliency detection and estimating depth maps simultaneously. Although the estimated depth awareness map is not as highly accurate as the captured one in~\figref{fig:motivation}, but focuses on more contrastive depth regions, which is more desirable for SOD tasks.
Second, the estimated depth awareness can be taken as an indicator to mine ambiguous regions. We calculate the logarithmic error between the estimated map and groundtruth depth to generate an adaptive weight map in~\figref{fig:motivation} (a). The network is further regularized to pay more attention to pixels with higher error-weighted responses, hence some semantic confusions can be gradually corrected.

The second challenge is to explore low-level cues for cross-modality interaction,~\ie, boundary awareness. Networks could be confused when encountering ambiguous regions in both RGB and depth data. These confusions would lead to fuzzy intermediate features when fusing the features from depth modal and RGB modal. Aggregating these features progressively may finally lead to inferior saliency results. Therefore, to make clear object boundaries for saliency detection, we introduce an edge attention module\footnote{We use the terms boundary and edge interchangeably for better notation.}
for cross-modality fusion in~\figref{fig:motivation} b). Features should be aware of sharp boundaries to enhance their confidence in ambiguous regions. In addition, unlike previous works using simple summation or multiplication~\cite{cheng2014depth,zhu2017innovative} for cross-modality fusions, we introduce depth features as attention weights to enhance the RGB features, finding the consensus salient regions in both modalities.

The third challenge relates to the fact that salient regions are invariant in multi-scale features. Due to the intrinsic flaw of CNN in perceiving multi-scale objects, previous works~\cite{chen2016attention} propose to re-scale the input image and ensemble the multi-scale output results. Besides, pyramid spatial pooling~\cite{zhao2017pyramid} offers another way by resampling the same feature layer into different scales. Despite the time-consuming issues, we found these techniques could not provide stable improvements in preferable saliency results.
To tackle the scale issue in the RGB-D SOD problem, we provide an innovative Gated Multi-Scale (GMS) predictor, which employs multi-scale predictors with a gated selection in~\figref{fig:motivation} c). We first re-sample the image at multiple scales and make each predictor identically map with one input scale. In the inference stage, these predictors are simultaneously activated and fused with the original branch to form a multi-scale output. Besides its notable improvements, our proposed GMS can be plug-and-play into different network architectures and introduces less additional computation cost.

\begin{figure*}
	\begin{center}
		%\fbox{\rule{0pt}{2in} \rule{.9\linewidth}{0pt}}
		\includegraphics[width=1\textwidth]{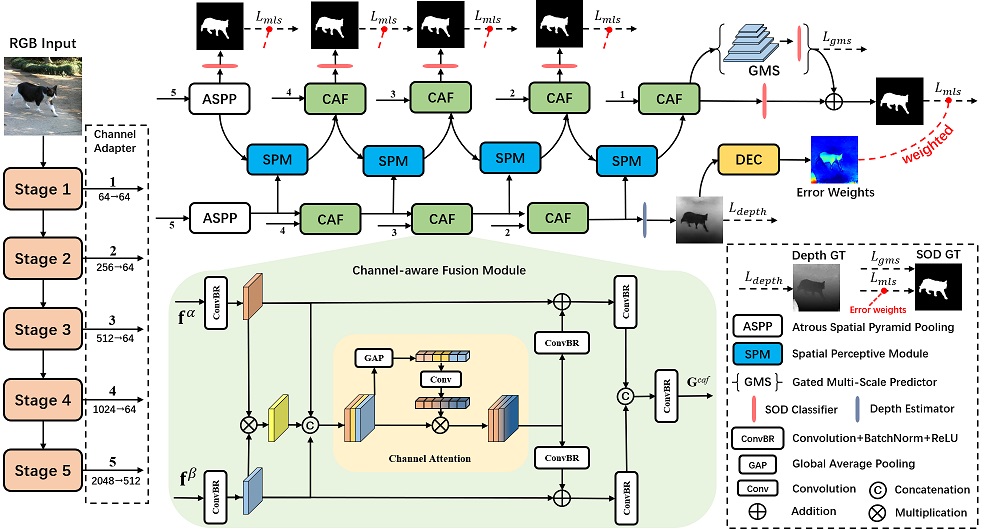}
		\caption{The overall architecture of our proposed Ubiquitous Target Awareness (UTA) network, which introduces three awareness into network embedding. 1) depth awareness: introducing depth supervision and depth error-weighted correction (DEC) module. 2) low-level cues awareness: spatial perceptive module (SPM) for cross-modality fusion with boundary supervision. 3) scale awareness: introducing a plug-and-play gated multi-scale (GMS) module. In each modality, a channel-aware fusion module (CAF) is proposed to select the channel-wise salient consensus features.
		}\label{fig:architecture}
	\end{center}
\end{figure*}

In this paper, we present a real-time Ubiquitous Target Awareness (UTA) network, leveraging the advantage of depth information, multi-scale, and low-level cues. As shown in~\figref{fig:architecture}, we first propose to adopt a depth-awareness constraint to regularize the features in different levels of the network stage while learning the object segmentation in the meantime. This forces the segmentation features to be aware of contrastive objects in the depth of field.
On the other hand, we utilize a depth error-weighted map to emphasize the saliency ambiguous regions,~\ie, objects salient in depth maps but not in RGB images. These regions are attached with more attention in the overall learning procedure for alleviating the object confusion and generating clear object shapes. Based on this awareness of depth modality, we propose a novel Spatial Perceptive Module (SPM) for cross-modal fusion, which is composed of two essential parts. The first depth spatial attention provides guidance to the RGB stream, focusing and selecting common features. And the second part regularizes the fused features to be aware of sharp boundaries, generating distinct salient regions. For intra-modality fusion strategies, we propose a selective Channel-Aware Fusion (CAF) module to strengthen salient consensus features. With all these informative modules combined, we finally propose a gated multi-scale predictor to handle the multi-scale object existences.

Contributions of this paper are summarized as follows:

1) We first set out a novel depth-aware setting for RGB-D salient object detection and propose a Ubiquitous Target Awareness network to solve this important problem. 2) We propose a depth awareness module to facilitate the understanding of saliency and design a depth-aware error-weighted loss to mine ambiguous pixels.  3) We propose a channel-aware fusion module to adaptively select cross-level features and a spatial perceptive module to capitalize on depth-aware and low-level cues for cross-modal fusion.
4) We propose an effective gated multi-scale predictor to further boost performance with the mutual complementation of multi-scale features.
5) We conduct extensive experiments on 10 benchmarks to demonstrate the superiority of our real-time framework in promoting RGB-D SOD tasks with only RGB inputs and validate its extensibility on RGB SOD tasks with estimated training depth.

The remainder of this paper is organized as follows:~\secref{sec:relatedwork} reviews related works and~\secref{sec:method} describes the proposed ubiquitous target awareness network for RGB-D salient object detection. Qualitative and quantitative experiments are reported in~\secref{sec:experiment}.~\secref{sec:conclusion} finally concludes this paper.

\section{Related Work}\label{sec:relatedwork}
 In this section, we first review the related works about the RGB-D salient object detection and analyze the differences of our work. As our proposed method can also adapt to handling the RGB scenarios, we briefly introduce the development of RGB SOD methods. At last, we review the recent works in the field of depth estimation.

\begin{figure}
	\begin{center}
		%\fbox{\rule{0pt}{2in} \rule{.9\linewidth}{0pt}}
		\includegraphics[width= \linewidth]{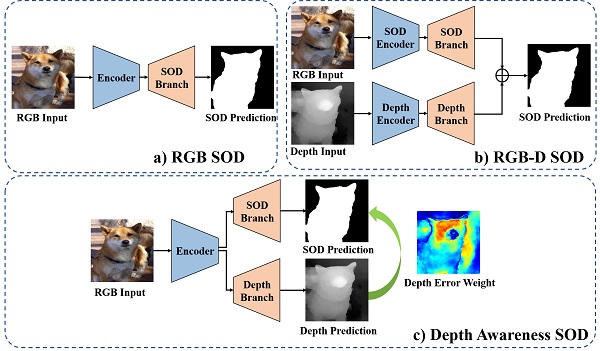}
		\caption{Different types of SOD architecture. a): Typical RGB SOD network architecture. b): Typical RGB-D SOD network architecture. c): Proposed Depth-awareness SOD network architecture.
		}\label{fig:introduction}
	\end{center}
\end{figure}

\textbf{RGB-D Salient Object Detection.}
It is shown that depth cues play an important and effective role in salient object detection. Besides methods~\cite{desingh2013depth,song2017depth} utilizing hand-crafted features, recent ideas~\cite{qu2017rgbd,chen2018progressively,chen2019multi,chen2019three,
zhang2020select,fan2019D3Net,chen2020dpanet,li2020asif,zhang2020uc,li2020rgb,li2020icnet,chen2020rgbd,Wei2020ECCV,oursMM,bai2021circular} propose to utilize deep models to learn prior knowledge from these two different modalities. As in~\figref{fig:introduction} b), existing RGB-D SOD models mainly rely on extracting salient features from RGB image and depth map respectively, and then fuse them in the early or late network stages~\cite{zhou2020rgb}.
 Following this trend, earlier work~\cite{peng2014rgbd} proposes to concatenate RGB-D pairs as 4-channel inputs for salient object detection. Considering the modality difference of depth and RGB data,
Han \etal~\cite{han2017cnns} propose a two-stream network to extract RGB features and depth features, and then fuse them with a combination layer. Piao \etal~\cite{piao2019depth} develop a two-stream network and fuse paired multi-level side-out features to refine the final salient object detection results. To extract the informative cues from both modalities, Li~\etal~\cite{li2020rgb} propose a spatial-aware selection and a channel-aware selection module to fuse the RGB and depth features.

However, directly fusing the depth cues and RGB information would lead to insufficient cross-modality understanding. Several works~\cite{li2021hierarchical,chen2018progressively,zhao2019contrast,zhao2019pyramid} propose to fuse these two modality encoders with a stage-wise or hierarchical manner. For example, Chen \etal~\cite{chen2018progressively} propose a progressive fusion strategy in a coarse-to-fine manner for sufficient information learning.  Zhao \etal~\cite{zhao2019pyramid} propose a fluid pyramid integration strategy to make full use of depth enhanced features.  Li~\etal~\cite{li2021hierarchical} tends to utilize the alternate interaction of different network stages for learning relations of different modalities.  Besides these fusion strategies, Chen~\cite{chen2020rgbd} propose to find a disentangled feature representation of each modality and learns to interact the same type of disentangled feature with others. Different from these aforementioned deep models, in this paper, we propose a new depth-awareness framework that does not rely on the depth data during the testing phase and learns the prior depth knowledge during training time.

\textbf{RGB Salient Object Detection.}
Conventional RGB SOD methods tends to hand-crafted features such as color constraints~\cite{cheng2014global}, texture~\cite{yan2013hierarchical} and local/global region contrast~\cite{klein2011center}.
Recently, CNN-based RGB SOD methods~\cite{liu2018picanet,li2019constrained,wu2019cascaded,li2021salient,zhao2019pyramid,su2019selectivity,qin2019basnet,F3Net,li2020parallel,ma2021pyramidal} have achieved impressive improvements over non-deep learning methods~\cite{cheng2014global,yan2013hierarchical,klein2011center}. Most of them are designed in end-to-end architectures in \figref{fig:introduction} a).
For example, Liu \etal~\cite{liu2018picanet} utilize pixel-wise contextual attention to select global and local contextual information. Zhao~\etal~\cite{zhao2019pyramid} propose a pyramid feature attention network, which adopts channel-wise attention and spatial attention to focus more on valuable features. Wei \etal~\cite{F3Net} propose a cross-feature module to fuse features of different levels and propose a boundary-sensitive loss for feature regularization.  Besides, some recent works~\cite{li2020personal,qin2019basnet,zhao2019egnet} also focus on the discovering the edge information when segmenting salient objects.

%Borji~\etal~\cite{borji2015salient} comprehensively review these methods for details with both deep learning and conventional hand-crafted techniques.
%Beyond these methods, recent research efforts pay attention to the boundary optimization of salient objects.
%Qin \etal~\cite{qin2019basnet} design a hybrid loss to focus on the boundary quality of salient objects.
%Wu \etal~\cite{wu2019cascaded} propose a coarse-to-fine aggregation framework, which discards low-level features to reduce the complexity.
%Su \etal~ \cite{su2019selectivity} propose a boundary-aware network to fuse the boundary and interior features with a compensation mechanism and an adaptive manner.

\textbf{Single Image Depth Estimation.}
 Methods of depth estimation can be divided into three groups: monocular video~\cite{wang2018learning}, stereo image pairs~\cite{garg2016unsupervised} and single image~\cite{eigen2014depth,eigen2015predicting,laina2016deeper,fu2018deep,yin2019enforcing}.  With the development of deep networks~\cite{simonyan2014very,he2016deep}, methods of depth information estimation~\cite{eigen2014depth,eigen2015predicting,laina2016deeper} or enhancement~\cite{he2021towards} has been boosted to a new accuracy level.
Eigen \etal~\cite{eigen2014depth,eigen2015predicting} propose a CNN-based framework for single image depth estimation, utilizing a stage-wise multi-scale network for further refinement.
As an important cue in computer vision tasks, recent works tend to utilize multi-task learning to joint depth estimation and other pixel-level prediction tasks, such as semantic segmentation~\cite{mousavian2016joint}, surface normal prediction~\cite{yin2019enforcing}.

%Laina \etal~\cite{laina2016deeper} introduce a fully convolutional architecture and design reverse Huber loss to the smoothness effect of L2 normalization.
%Taking a single image as input is the hardest case because there is less geometric information.
%Fu~\etal~\cite{fu2018deep} propose a spacing-increasing discretization strategy to discretize depth and recast depth estimation as an ordinal regression problem. Yin \etal~\cite{yin2019enforcing} propose a global geometric constraint to improve depth estimation accuracy.

\section{METHODOLOGY}\label{sec:method}
\subsection{Overview}
In this section, we introduce a novel Ubiquitous Target Awareness (UTA) network for RGB-D salient object detection. The main idea of our framework is to leverage depth, multi-scale, and low-level spatial constraints to regularize the feature learning process. These regularizations are naturally embedded in one unified network, endowing the learned features to be aware of specific representations.

\begin{figure}[t]
	\begin{center}
		%\fbox{\rule{0pt}{2in} \rule{.9\linewidth}{0pt}}
		\includegraphics[width=\linewidth]{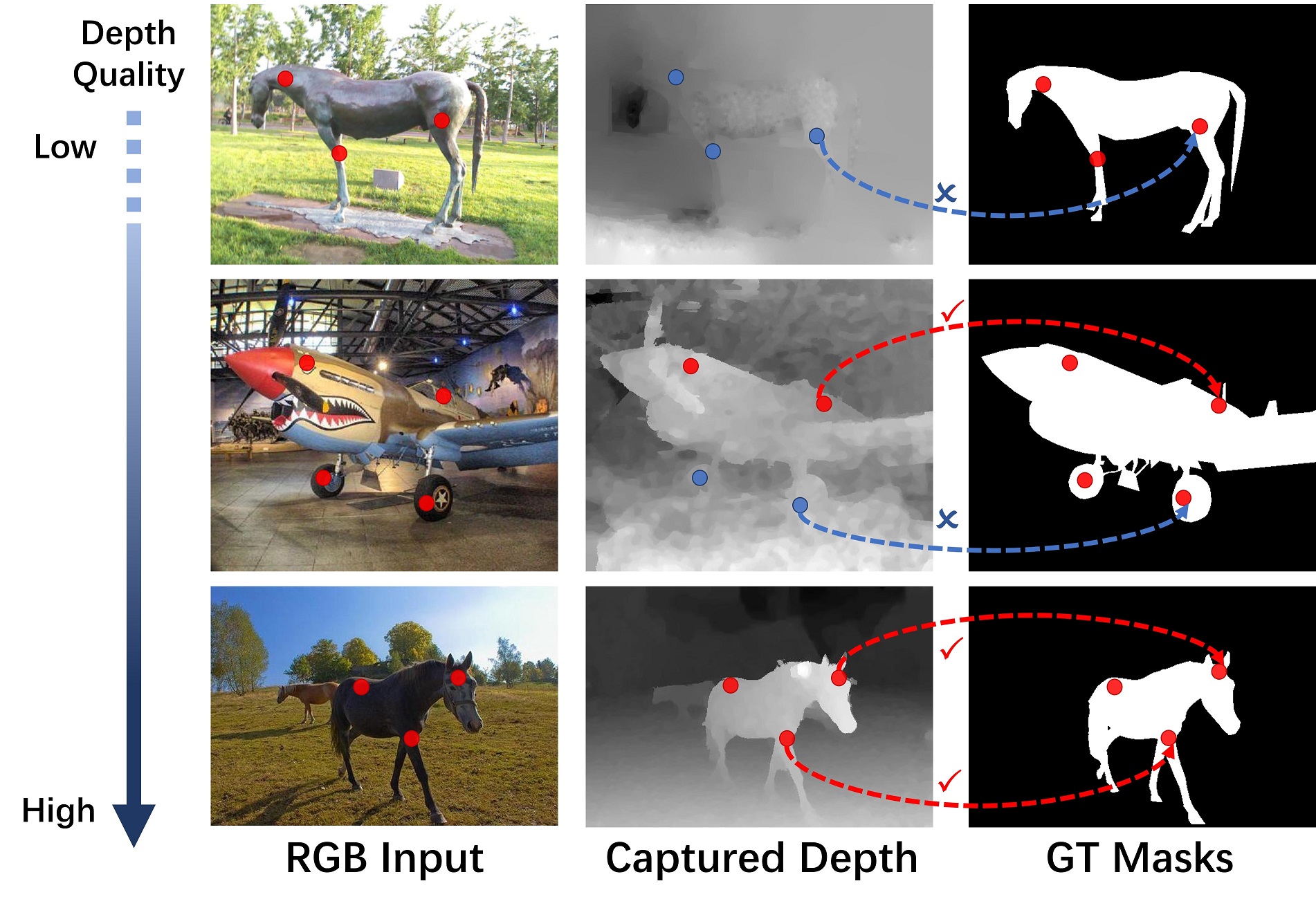}
		\caption{The motivation of depth-awareness SOD. Depth maps of high quality can provide clear low-level cues for SOD learning while the low-quality ones usually introduce noisy information. The red and blue dotted lines indicate the positive and negative effects on the learning process respectively.
		}\label{fig:depth_dis}
	\end{center}
\end{figure}

\textbf{Motivation of Depth-awareness SOD.} As mentioned above, conventional RGB-D methods adopt an RGB encoder for the three-channel RGB input while using a depth encoder to encode the captured depth image. Then the extracted features are fused together to get the final output. However, this learning manner leaves a major concern: is depth information always useful for salient object detection?

In~\figref{fig:depth_dis}, the quality of captured depth maps varies greatly under different circumstances. These phenomena are inevitable considering the various imaging conditions. From the low quality to high quality of depth images, it would lead to different learning manners of salient object detection tasks. The high-quality depth maps usually provide explicit low-level cues, involving boundaries, textures, and color contrasts. However, the low-quality depth maps in the first row of~\figref{fig:depth_dis} provide noisy information for salient object learning. Existing works utilize the depth data in a consistent manner and regard it as positive information under all circumstances. However, when facing the depth data of low-quality, these methods would incorporate noisy inputs and harm the SOD learning performance.

To solve this important problem, we explore the depth data as a learnable prior knowledge rather than directly encoding the raw depth data. This designment shows following advantages:
 1) The estimated depth data plays a stable and positive role in complementing the RGB branch and their quality would not be restricted by the imaging condition.
2) We propose to correct the errors in the depth-saliency ambiguous regions, where should be attach more importance during the learning process.
3) Our framework takes only RGB data as input, while not relies on the capture of depth data. This advantage greatly extends our framework into some extreme scenarios without clear depth data, or in single-modal RGB applications.

\textbf{Architecture of Depth-awareness SOD.} Let $\br{I}_{RGB}$, $\br{I}_{D}$ and $\br{M}$ denote the input RGB image, depth data and SOD mask respectively. $\mc{E}$ and $\mc{D}$ denotes the encoder and decoder network in SOD architecture.  Conventional RGB SOD methods in~\figref{fig:introduction} a) takes only RGB images as input,~\ie, $\br{M}=\mc{D}(\mc{E}(\br{I}_{RGB}))$. With depth maps as auxiliary inputs in~\figref{fig:introduction} b), the overall framework requires two independent encoders to extract the depth and RGB features separately, which main computation costs are usually lied in. Methods of this category~\cite{fan2020bbs,liu2020learning} can be similarly denoted as $\br{M}=\mc{D}_R(\mc{E}_{R}(\br{I}_{RGB})) \biguplus \mc{D}_D(\mc{E}_{D}(\br{I}_{D}))$, where $\biguplus$ denotes the cross modal fusion strategies.
Moreover, the depth and RGB encoders are separately trained and the relationships between these multi-modal data are not fully explored.
Taking only RGB inputs as well as saving the computation costs, the depth-aware salient object detection in~\figref{fig:introduction} c) provides us a new perspective to utilize the depth data in this segmentation task. In the testing phase, the network only takes the RGB as input and the object segmentation results are regularized by the depth-awareness constraints in the training phase. This testing process can be represented as $\br{M}=\mc{D}_R(\mc{E}(\br{I}_{RGB})) \biguplus \mc{D}_D(\mc{E}(\br{I}_{RGB}))$, where $\mc{D}_D$ is supervised by captured depth groundtruth.

In this framework, after the weight-sharing encoder network, we propose a depth-awareness branch to predict the depth information in~\figref{fig:architecture}. Thus depth features in this auxiliary branch benefiting SOD in two aspects: 1) passing into the depth-error correction (DEC) to help mine ambiguous regions; 2) serving as depth modality data to enhance and fuse with the original RGB branch. For the cross-modality fusion, we propose a spatial perceptive module (SPM) to force SOD features to be aware of depth data and boundary cues. For the fusion of intra-modality, we present a channel-aware fusion module, finding salient consensus features and alleviating the noisy ones.
After that, we propose a plug-and-play gated multi-scale predictor (GMS) to perceive the multi-scale object existences. Benefiting from the ubiquitous target awareness, our framework does not rely on depth data or edge information in the testing phase, generating high-quality segmentation masks with intrinsic features.

\subsection{Depth-awareness Constraint}
What role does depth information play in salient object detection? To answer this question, in this paper, we propose an innovative depth-awareness constraint from two complementary aspects,~\ie, multi-level depth awareness, and depth error-weighted correction. These two aspects work collaboratively to regularize the salient features being aware of contrastive depth regions and contextual salient confusions, which facilitate the segmentation process in different learning stages.

\textbf{Multi-level Depth Awareness.} As discussed in previous works~\cite{zhang2018progressive,fan2020bbs,chen2018progressively}, the key issue in salient object detection lies in the utilization of multi-level features in different network stages. Besides the aggregation strategy, the other exploitation is to regularize the features focusing on meaningful regions, which provides useful contextual information before aggregation. Taking the advantages of depth information and the hierarchical network architecture, we force the segmentation features to focus on depth regions in different network learning stages, which is elaborated in~\figref{fig:architecture}. This means in each network learning stage, features should be aware of the object information as well as depth-contrastive regions. We use an additional depth branch to regress the groundtruth depth. With this collaborative learning of SOD and depth regression, we further fuse these two branches to refine the salient object detection results (see~\figref{fig:architecture}), which builds strong correlations between these two different types of features. Notably, this refinement strategy can also be well handled by the spatial perceptive module (SPM), with additional boundary supervision at multiple levels. As a result, the salient features stand as a predominant place in the final optimization and the depth maps become leading guidance.

\textbf{Depth Error-weighted Correction.}
To make thorough exploitation of depth information, we further propose a depth error-weighted correction (DEC) to regularize hard pixels with higher weights where the predicted depth makes mistakes. As it stands, the network itself naturally tends to be highly responded to the salient regions and then form a holistic salient object. However, this would guide the predicted depth features focusing on salient regions, resulting in a severe misalignment between the predicted depth and groundtruth data. Remarkably, the error regions where the predicted depth makes mistakes are usually the semantic ambiguous regions, which we need to pay more attention to in the learning process.

 \begin{figure}[t]
 	\begin{center}
 		%\fbox{\rule{0pt}{2in} \rule{.9\linewidth}{0pt}}
 		\includegraphics[width= \linewidth]{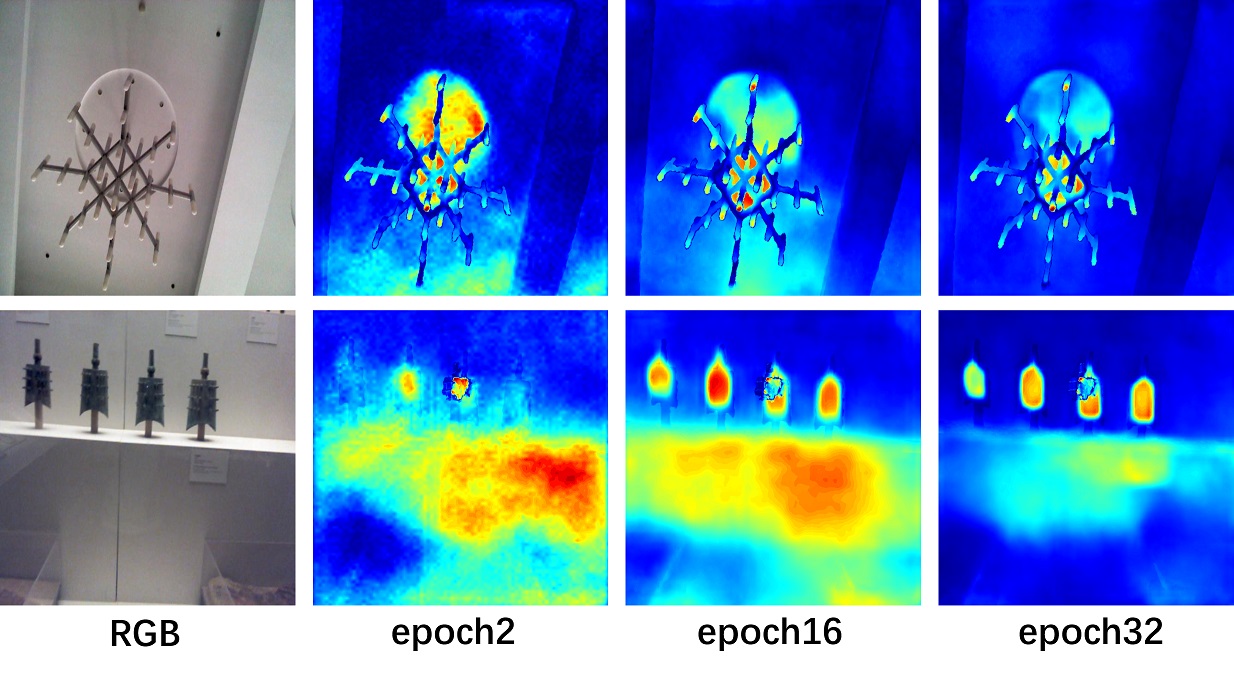}
 		\caption{Qualitative visualization of depth error weights in different training epochs. The ambiguous error regions can be iteratively optimized during the training process.
 		}\label{fig:weights}
 	\end{center}
 \end{figure}

In order to solve this misalignment as well as to exploit it, we thus introduce a logarithmic depth error weight. Let $p^d$ and $y^d$ be the predicted depth and  depth groundtruth (GT) respectively, the error weight $\br{e}_{ij}$ of each pixel has the form:
\begin{equation} \label{eq:dew}
\br{e}_{ij}=\frac{\sum_{i=1}^{h}\sum_{j=1}^{w}(\log p_{ij}^d-\log y_{ij}^d)}{\sum_{i=1}^{h}\sum_{j=1}^{w} \max(\log p^d-\log y^d)},
\end{equation}
where $w$ and $h$ are the width and height of the error window, which represents the error of central pixel with the mean value of local regions.
In this way, the ambiguous pixels are treated with more attention in the early learning phase. As the optimization goes through, the regularized features become depth-aware, and errors are progressively corrected. This learning progress is exhibited in~\figref{fig:weights}, where the highly-responded regions in the error map shrink along with the learning stage. This verifies that the final optimized features are aware of depth information and better at handling semantic confusions.

\subsection{Channel-aware Cross-level Interaction} \label{sec:CAF}
The crucial problem in salient object detection is to select the most discriminative features and to pass them in a coarse-to-fine scheme. However, aggregating features from different levels in an encoder-decoder fashion usually leads to missing details or introduces ambiguous features, which jointly lead to a bad network optimization. Besides, high-level features show a strong ability to capture salient regions but fail to generate finer boundaries, while low-level features contain boundary and detailed cues but also more noisy features.

Toward this end, we propose a novel Channel-Aware Fusion module (CAF), which adaptively selects the discriminative features for object understanding.
The proposed CAF shows some meaningful designs, which are illustrated in~\figref{fig:architecture}. First, given two types of source feature $\br{f}^\alpha, \br{f}^\beta \in \mathbb{R}^{W'\times H'\times C'}$, we use pixel-wise multiplication to enhance the common pixels in feature maps, while alleviating the ambiguous ones. The enhanced features are then concatenated with the transformed features with a lightweight encoder $\mc{E}(\cdot)$. This operation can be formally represented as:
\begin{equation} \label{eq:eq feature fusion process1}
\br{f}^{ca} = \mc{E}_\alpha(\br{f}^\alpha) \copyright \mc{E}_\beta(\br{f}^\beta) \copyright (\mc{E}_\alpha(\br{f}^\alpha) \otimes \mc{E}_\beta(\br{f}^\beta)),
\end{equation}
where $\copyright$ and $\otimes$ denote the feature concatenation operation and pixel-wise multiplication respectively. Each encoder $\mc{E}_{\{\alpha,\beta\}}$ is typically composed of a $3\times 3$ convolutional layer followed by a Batch Normalization and a ReLU activation.
Especially, when aggregating the multi-level features, the features $\br{f}^\alpha$ and $\br{f}^\beta$ are first upsampled to the same scale, which is omitted for better view in~\figref{fig:architecture}.

After obtaining rich feature $\br{f}^{ca}\in \mathbb{R}^{W'\times H'\times 3C'}$ by Eq.~\equref{eq:eq feature fusion process1}, the second main concern is how to select the most relevant features that are highly-responded in the segmentation target. Inspired by the channel-attention mechanism~\cite{hu2018squeeze,chen2017sca}, we thus propose to use global features for a contextual understanding of the attention weights. The $\br{f}^{c}$ is squeezed with a global average pooling, followed by a sigmoid normalization $\sigma$, and is then transformed as the vector shape to align the channel dimension with the original feature.
This serialized operation has the form:
\begin{equation}\label{eq:eq feature fusion process2}
\br{a}=\frac{1}{W'\times H'} \sum_{i=1}^{W'}\sum_{j=1}^{H'}\br{f}^{ca}_{i,j},
\end{equation}
\begin{equation} \label{eq:eq feature fusion process3}
\br{u}_{i,j,k} = \br{f}^{ca}_{i,j,k} \otimes \sigma(\varphi(\br{a}_{k})).
\end{equation}
$\varphi$ is a linear transformation to reorganize the pooling features and $\br{u}$ denotes the learned attention weighted features. Therefore features relevant to the salient target could be prominent in each group of source features $\br{f}^\alpha$ and $\br{f}^\beta$. This can be achieved by a channel-aware attention mechanism:
\begin{equation} \label{eq:eq feature fusion process3}
\br{G}^{caf} = \mc{R}_g (\mc{E}_{v1}(\mc{E}_{\alpha}(\br{f}^\alpha) \oplus \mc{R}_{u1}(\br{u}))
\copyright
\mc{E}_{v2}(\mc{E}_{\beta}(\br{f}^\beta) \oplus \mc{R}_{u2}(\br{u}))),
\end{equation}
where $\mc{R}_{\{u1,u2,g\}}$ denotes the typical shape-preserving decoder and $\mc{E}_{\{v1,v2\}}$ denotes the dimensional reduction decoder, forming a same shape as original inputs. Hence the relevant features to target object can be enhanced in the final output $\br{G}^{caf}$.
In addition, to implement the whole framework in a lightweight trend, the channel dimension $C'$ is empirically set as $64$ to achieve state-of-the-art performance.

\begin{figure}[]
	\begin{center}
		%\fbox{\rule{0pt}{2in} \rule{.9\linewidth}{0pt}}
		\includegraphics[width=\linewidth]{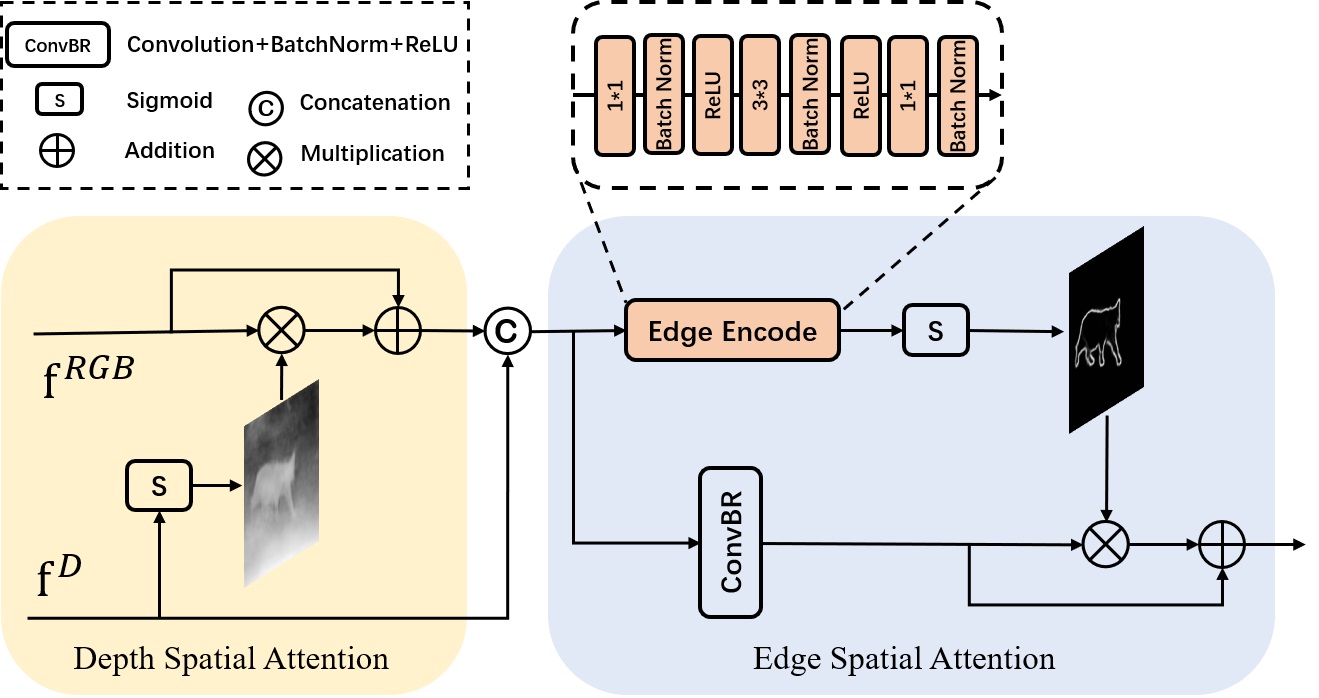}
		\caption{Illustration of the proposed Spatial Perceptive Module (SPM) for cross-modal interaction. Features from the RGB stream are first enhanced with depth attention to find the salient coherent regions and then fed into an edge spatial attention module to incorporate the low-level cues in the fusion process.
		}\label{fig:SPM}
	\end{center}
\end{figure}

\subsection{Spatial-aware Cross-modal Interaction}
In this subsection, we propose a spatial perceptive module (SPM) to handle the cross-modal interaction of depth data and RGB data.
Unlike the above-mentioned fusion strategies, we propose to enhance the feature representation using two attention modules,~\ie, depth spatial attention and edge spatial attention.
The cross-modal fusion first aims to find the saliency consensus in both depth and RGB data. As complementation to RGB data, we leverage the normalized depth attention map as guidance to help the saliency inference. With given RGB features $\br{f}^{RGB}$ and depth features $\br{f}^{D}$, this operation could be formally represented as:
\begin{equation} \label{eq:eq cmf 1}
\begin{split}
\br{f}^{dsa}= (\br{f}^{RGB} \otimes \sigma(\br{f}^{D}) \oplus \br{f}^{RGB})\copyright \br{f}^{D},
\end{split}
\end{equation}
where $\br{f}^{dsa}$ denotes the depth-enhanced feature with spatial-aware depth attentions, and $\sigma(\cdot)$ is the sigmoid normalization.
On the other hand, although depth data provide high-level saliency guidance for the fused features, low-level cues are usually neglected during the message passing. In addition, depth data are sometimes captured in low-quality, progressively fusion with multi-stage decoders would aggravate this phenomenon. Thus we introduce the boundary information as the low-level guidance to form a clear object shape. To explicitly embed the edge features into the network, we use a feature encoder $\mc{E}_{edge}$ consisting of $1\times1$, $3\times3$ and $1\times1$ convolutional blocks with BatchNorm and ReLU supervised by edge masks. The salient boundaries are generated by canny edge detectors and expanded to a width of five pixels.

With the encoded edge features, we then pass them with a sigmoid function to generate normalized attention maps. The depth-enhanced features $\br{f}^{dsa}$ pass by a convolutional block $\mc{R}_{s}$ to reduce its channels and conduct a pixel-wise spatial attention mechanism:
\begin{equation} \label{eq:eq cmf 2}
\br{G}^{spm}_{i,j} = (\mc{R}_{s}(\br{f}^{dsa}_{i,j}) \otimes \sigma(\mc{E}_{edge}(\br{f}^{dsa}_{i,j})))\oplus \mc{R}_{s}(\br{f}^{dsa}_{i,j}).
\end{equation}
The proposed SPM modules are embedded in multiple cross-modal fusion stages. Progressively employing this module helps keep the low-level details in the network learning stage, which is beneficial to generate sharp masks.

%How to fuse cross-modal information is an unique problem in RGB-D salient object detection. However, directly fuse these two different modal features with addition or concatenation \etal usually lead to lower performance, which may be caused by two aspects, the loss of each modal's distinct features caused by the inappropriate fusion manner and the lack of edge details in depth-modal features. To solve these two issues, we propose a simple yet effective cross-modal fusion module (CMF), which utilizes depth features as guidance to enhance sod features and makes up edge details with spatial attention mechanism.

%To enhance sod features with depth guidance, we extract the geometry salient regions in depth features to enhance sod features by multiplication and addition operation. We retain depth details by concatenation operation. It can be formally represented as:

\begin{figure}[t]
	\begin{center}
		%\fbox{\rule{0pt}{2in} \rule{.9\linewidth}{0pt}}
		\includegraphics[width=\linewidth]{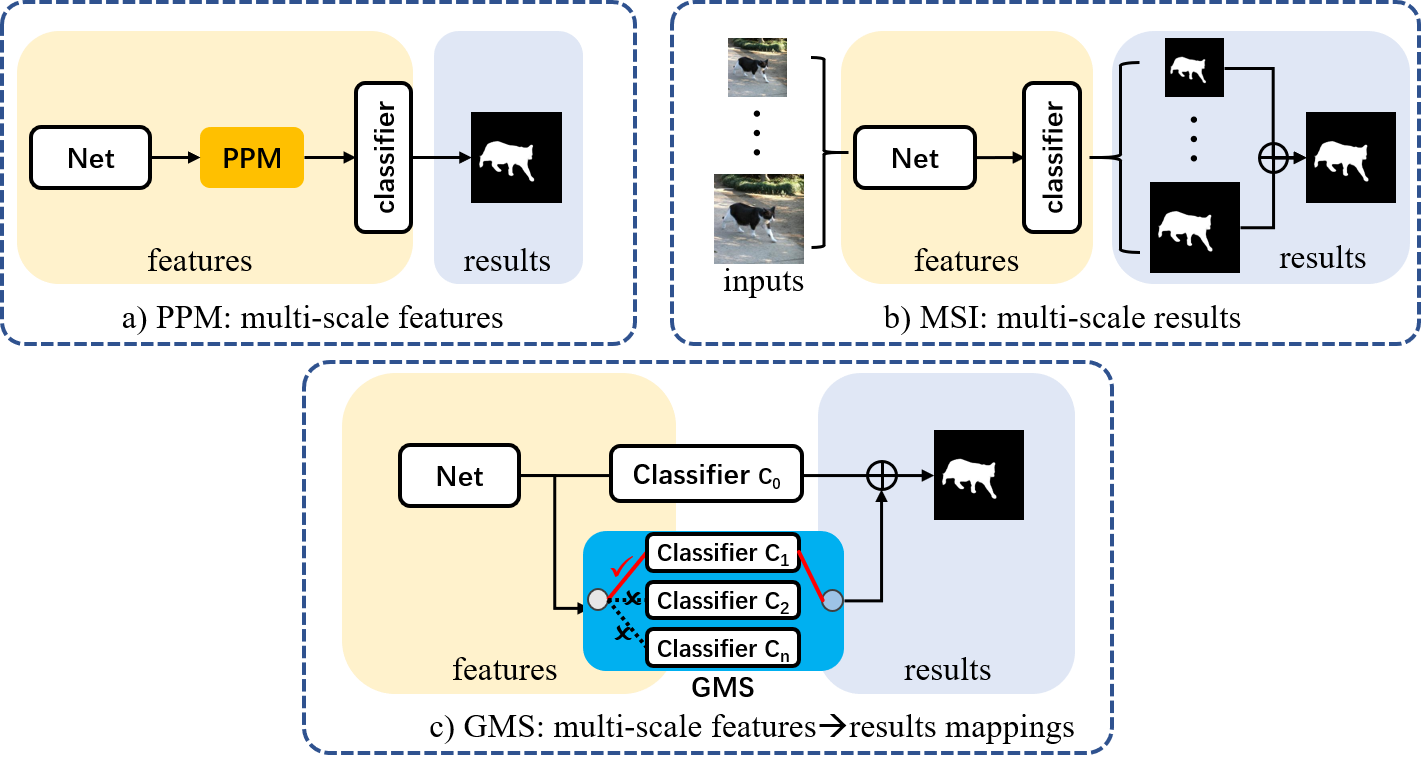}
		\caption{Comparisons of three multi-scale strategies. a) PPM~\cite{zhao2017pyramid}: aggregating multi-scale features to the new feature.  b) MSI~\cite{chen2016attention}: aggregating multi-scale results to the new result. c) proposed GMS: naturally embedded in network prediction and transforming features to results.}\label{fig:GMS difference}
	\end{center}
\end{figure}

\begin{figure}[t]
	\begin{center}
		%\fbox{\rule{0pt}{2in} \rule{.9\linewidth}{0pt}}
		\includegraphics[width=0.9\linewidth]{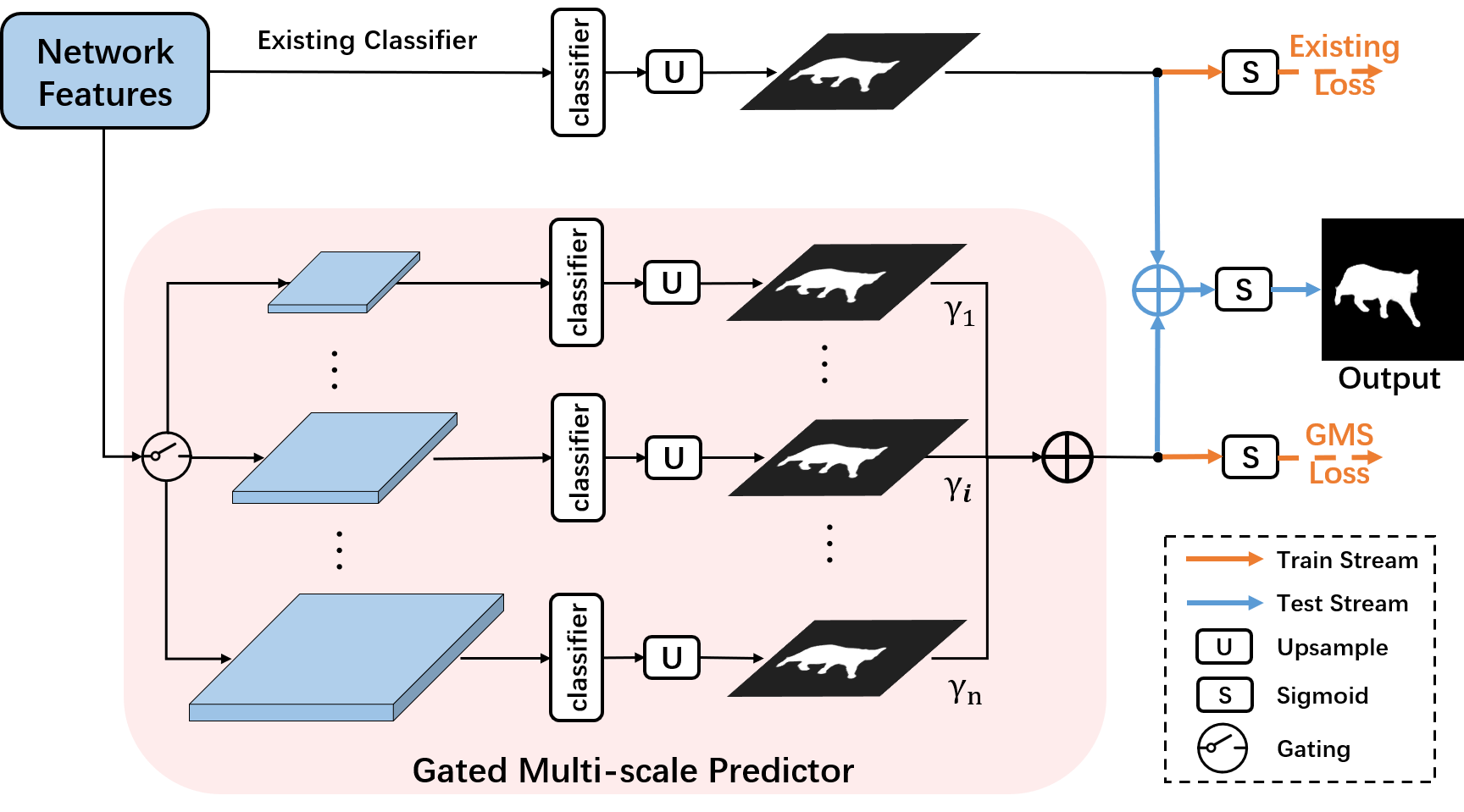}
	\caption{Illustration of the proposed Gated Multi-Scale (GMS) predictor. Multiple individual classifiers with $1\times 1$ linear transformations are adopted to grasp the scale-related features.  The existing loss in our manuscript denotes the previously defined loss in Eqn. (16), ~\ie, $\mc{L}_{depth}$, $\mc{L}_{edge}$, $\mc{L}_{mls}$.}\label{fig:GMS}
	\end{center}
\end{figure}

\subsection{Gated Multi-Scale Predictor}\label{sec:gms}
There is a natural defect in deep CNNs to perceive multi-scale features, due to the fixed down-sample and up-sample operations. In particular we consider two predominant approaches in tackling this issue: 1) multi-scale inference (MSI)~\cite{chen2016attention}: ensembling outputs of multi-scale inputs,~\eg, rescaled from $0.5$ to $2.0$ during inference; 2) pyramid pooling module (PPM) \cite{zhao2017pyramid,zhao2019pyramid}: pooling the feature map to multiple scales of the same layer. These two operations are precisely exhibited in~\figref{fig:GMS difference}.

Here we elaborate on these methods in tackling SOD tasks:
multi-scale inference techniques suppose that objects should be relatively salient despite their scales. However, once the network is optimized, the understanding of salient objects is usually fixed. This relative saliency relationship is hard to obtain. Besides MSI, the pyramid pooling module (PPM)~\cite{zhao2017pyramid} introduces variant scales by resizing the feature map of the same layer. By introducing additional computation costs, the reception field of the same pixel could be enlarged.
%However, this operation introduces additional learning parameters for fusion and hyper-parameters for pooling size, which sometimes are hard to optimize. We experimentally found these techniques could not provide stable improvements on preferable salient detection results.

Different from these above methods, we propose a new gated multi-scale predictor in~\figref{fig:GMS}, which provides several technical insights in salient object detection. We observe that it is the last SOD classification layer that affects the result mostly. Before that, rich features of multiple scales are contained in a high dimension of $\mathbb{R}^{W\times H\times C}$. The final classification layer aims to excavate the most reliable SOD map $\mathbb{R}^{W\times H\times 1}$ from these features, which can be regarded as a mapping function $\mc{F}$. However, this operation is conducted into a single-scale trend. When object scales change, this learned single classifier conducts the same mapping function $\mc{F}$, leading to inferior results.

Essentially speaking, the final mapping function $\mc{F}$ would significantly affect the quality of salient object detection. The mapping function is composed of two operations,~\ie, a rescaling operation and a $3\times 3$ convolutional layer.
Thus we propose to use multi-scale predictors with a set of mapping functions,~\ie, $\mc{F}^1,\mc{F}^2,\cdots,\mc{F}^n$ of $n$ scales. In the training phase, we rescale the input image to different sizes,~\ie,  [224, 256, 288, 320, 352]. When images of one scale are fed into the network, only one scale predictor and the backbone classifier are activated, which are controlled by a gating function.  In the testing phase, only one scale of input is fed into the network, and all mapping functions are activated to perceive a multi-scale saliency understanding.
The output features are finally weighted for summation in~\figref{fig:GMS} and then fused with the existing classifier. The weights $\gamma$ is set as [0.25, 0.25, 0.25, 0.25, 1] for five GMS branches respectively.
Moreover, our proposed GMS is a plug-and-play module in parallel with the existing SOD classifier, which is lightweight and can further boost the performance based on preferable results.

\subsection{Learning Objective}
As a typical SOD task, our framework is first supervised with the SOD masks, with a typical Binary Cross-Entropy (BCE) loss. Let $p^s, y^s \in \mathbb{R}^{W\times H\times 1}$ be the predicted salient mask and corresponding groundtruth, the BCE loss has the form:
\begin{equation}\label{eq:bce}
\begin{split}
\mc{L}_{bce} = -\sum_{i=1}^{H}\sum_{j=1}^{W}y^s_{ij}\log(p^s_{ij}).
\end{split}
\end{equation}

However, the BCE loss usually leads to noisy predictions which do not form a holistic object. To make the salient object with clear boundaries, we adopt an Intersection over Union (IoU) measurement~\cite{F3Net, qin2019basnet} as the auxiliary loss:
\begin{equation} \label{eq:iou}
\begin{split}
\mc{L}_{iou} = 1-\frac{\sum_{i=1}^{H}\sum_{j=1}^{W} (y^s_{ij}\times p^s_{ij})+1}{\sum_{i=1}^{H}\sum_{j=1}^{W}(y^s_{ij}+p^s_{ij}-y^s_{ij}\times p^s_{ij})+1}.
\end{split}
\end{equation}

Beyond the SOD supervision, our learning objective is supervised by multiple constraints, as in~\figref{fig:architecture}, the depth awareness module, the error-weighted correction, the auxiliary edge supervision for cross-modality fusion, and the scale-awareness constraints.
For the first depth-awareness constraint, we adopt the logarithmic Mean Square Error (logMSE) for supervision~\cite{eigen2014depth,eigen2015predicting} to generate smooth depth maps, and meanwhile providing the error weights:
\begin{equation} \label{eq:logmse}
\begin{split}
\mc{L}_{depth} = \frac{1}{W\times H}\sum_{i=1}^{H}\sum_{j=1}^{W} || \log y^d_{ij} - \log p^d_{ij} ||^2_2.
\end{split}
\end{equation}

For edge encoders in the spatial perceptive module, we adopt a BCE loss to predict edges as a binary classification task:
\begin{equation}\label{eq:edge bce}
\begin{split}
\mc{L}_{bce}^{edge} = -\sum_{i=1}^{H}\sum_{j=1}^{W}y^e_{ij}\log(p^e_{ij}),
\end{split}
\end{equation}
where $y^e_{ij}$ and $p^e_{ij}$ denote the edge labels and predicted edges respectively. Then with four SPM modules, the final summation loss $\mc{L}_{edge}$ can be represented as:
\begin{equation}\label{eq:edge bce sum}
\begin{split}
\mc{L}_{edge} = \sum_{e=1}^{4}\mc{L}_{bce}^{edge}.
\end{split}
\end{equation}

For the error-weighted correction, we adopt an error-weighted BCE loss to attach more importance to the wrongly-predicted pixels:
\begin{equation} \label{eq:wbce}
\mc{L}_{dec} = \frac{-\sum_{i=1}^{H}\sum_{j=1}^{W}\br{e}_{ij}\times y^s_{ij}\log(p^s_{ij})}{\sum_{i=1}^{H}\sum_{j=1}^{W}\br{e}_{ij}}.
\end{equation}
This error loss $\mc{L}_{dec}$ adopts the same SOD groundtruth masks $y^s_{ij}$. To implement the multi-level supervision in a unified framework, the multi-level loss $\mc{L}_{mls}$ can be formulated as:
\begin{equation} \label{eq: MLS loss}
\begin{split}
\mc{L}_{mls} = \sum_{i=1}^{S}\lambda_i(\mc{L}_{bce}+\mc{L}_{iou}+\mc{L}_{dec}),
\end{split}
\end{equation}
where $\lambda_i$ denotes the weights of multi-level losses and $S$ is set as 5 representing five stages in ResNet. Here we follow GCPANet~\cite{chen2020global} and set $\lambda$ as [1, 0.8, 0.6, 0.4, 0.2].

For the gated multi-scale predictor of the final $S$ stages, we adopt an auxiliary loss $\mc{L}_{gms}$ to learn scale-aware information:
\begin{equation} \label{eq: GMS loss}
\begin{split}
\mc{L}_{gms} =\mc{L}_{bce}+\mc{L}_{iou}+\mc{L}_{dec}.
\end{split}
\end{equation}
With the above regularizations combined, the overall loss function $\mc{L}_{sum}$ can be formulated as:
\begin{equation} \label{eq:loss}
\begin{split}
\mc{L}_{sum} = \mc{L}_{depth}+\mc{L}_{edge}+\mc{L}_{mls}+\mc{L}_{gms}.
\end{split}
\end{equation}

\linespread{1.2}
\begin{table*}[t]
    \centering{
	\caption{Performance comparison with 12 state-of-the-art RGB-D SOD methods on five benchmarks. Smaller $MAE$, larger $F_\beta^{max}$, $F_\beta^{mean}$ and $F_\beta^{w}$ indicates better performance. Results ranked in the first and second places are highlighted in bold and underlined.}
	\label{table: rgbd benchmark}
 	\resizebox{\textwidth}{!}{
	\renewcommand\tabcolsep{1.0pt}
	\begin{tabular}{l|c|cccc|cccc|cccc|cccc|cccc}
		\hline
		\multicolumn{1}{c|}{\multirow{2}{*}{METHODS}} &
		\multirow{2}{*}{FPS} &
		\multicolumn{4}{c|}{NJUD-TE} &
		\multicolumn{4}{c|}{NLPR-TE} &
		\multicolumn{4}{c|}{STEREO} &
		\multicolumn{4}{c|}{DES} &
		\multicolumn{4}{c}{SIP} \\ %\cline{2-21}
		\multicolumn{1}{c|}{} & & $F_\beta^{max}$ & $F_\beta^{mean}$ & $F_\beta^{w}$ & $MAE$ & $F_\beta^{max}$   &  $F_\beta^{mean}$    & $F_\beta^{w}$ & $MAE$   & $F_\beta^{max}$ & $F_\beta^{mean}$ & $F_\beta^{w}$ & $MAE$ & $F_\beta^{max}$ & $F_\beta^{mean}$ & $F_\beta^{w}$ & $MAE$ & $F_\beta^{max}$ & $F_\beta^{mean}$ & $F_\beta^{w}$ & $MAE$ \\ \hline
		DF~\cite{qu2017rgbd}                    & 0.1 & .804 & .744 & .545 & .141 & .778 & .682 & .516 & .085 & .757 & .616 & .549 & .141 & .766 & .566 & .392 & .093 & .704 & .673 & .406 & .185 \\
% 		AFNet~\cite{wang2019adaptive}                 & .775 & .764 & .100 & .772 & .771 & .755 & .058 & .799 & .823 & .806 & .075 & .825 & .728 & .713 & .068 & .770 & .687 & .672 & .118 & .714 \\
		CTMF~\cite{han2017cnns}                & 1.6 & .845 & .788 & .720 & .085 & .825 & .723 & .679 & .056 & .831 & .786 & .698 & .086 & .844 & .765 & .687 & .055 & .720 & .684 & .535 & .139 \\
		MMCI~\cite{chen2019multi}                & 19 & .852 & .813 & .739 & .079 & .815 & .729 & .676 & .059 & .863 & .812 & .760 & .068 & .822 & .750 & .650 & .065 & .840 & .795 & .712 & .086 \\
		PCF~\cite{chen2018progressively}                 & 15 & .872 & .844 & .803 & .059 & .841 & .794 & .762 & .044 & .860 & .845 & .778 & .064 & .804 & .763 & .711 & .049 & .861 & .825 & .768 & .071 \\
		TANet~\cite{chen2019three}               & 14 & .874 & .844 & .805 & .060 & .863 & .796 & .780 & .041 & .861 & .828 & .787 & .060 & .827 & .795 & .740 & .046 & .851 & .809 & .748 & .075 \\
		CPFP\cite{zhao2019contrast}                & 7 & .876 & .850 & .828 & .053 & .869 & .840 & .807 & .036 & .874 & .842 & .817 & .051 & .838 & .815 & .787 & .038 & .870 & .819 & .788 & .064 \\
		DMRA~\cite{piao2019depth}                & 10 & .886 & .872 & .847 & .051 & .879 & .855 & .840 & .031 & .868 & .847 & .647 & .066 & .888 & .857 & .843 & .030 & .847 & .815 & .734 & .088 \\
		D3Net~\cite{fan2019D3Net}                 & 20 & .889 & .860 & .833 & .051 & .885 & .853 & .826 & .030 & .881 & .844 & .815 & .054 & .885 & .859 & .831 & .030 & .847 & .818 & .793 & .058 \\
		SSF~\cite{zhang2020select}  & 32 & \underline{.911} & .886 & .871 & .043 & .912 & .873 & .867 & .026 & .902 & .880 & .862 & .044 & .912 & .882 & .852 & .026 & - & - & - & - \\
		%UCNet~\cite{zhang2020uc}  & 16 & .908 & .889 & .868 & .043 & .915 & .890 & .878 & .025 & .908 & .885 & .867 & .039 & .937 & .908 & .907 & .019 & .896 & .868 & .836 & .051 \\
		CoNet~\cite{Wei2020ECCV}  & \underline{34} & .902 & .872 & .849 & .047 & .898 & .848 & .842 & .031 & .912 & .885 & \underline{.871} & .040 & .916 & .862 & .848 & .027 & .883 & .842 & .803 & .063 \\
		DCMF~\cite{chen2020rgbd}  & - & .898 & .859 & .831 & .047 & .895 & .839 & .828 & .031 & .887 & .841 & .811 & .047 & .879 & .812 & .780 & .027 & .866 & .819 & .780 & .068 \\
		ICNet~\cite{li2020icnet} & 13 & .891 & .869 & .843 & .052 & .908 & .884 & .864 & .028 & .898 & .870 & .844 & .045 & .913 & \textbf{.893} & \textbf{.867} & .027 & .857 & .834 & .791 & .069 \\
		\hline
		Ours (DASNet) \cite{oursMM} & 33 &
		\underline{.911} &
		\underline{.894} &
	    \underline{.872} &
		\underline{.042} &
		\underline{.929} &
		\underline{.907} &
		\underline{.892} &
		\underline{.021} &
		\underline{.915} &
		\underline{.894} &
		.870 &
		\underline{.037} &
		\textbf{.928} &
		\underline{.892} &
		\textbf{.867} &
		\textbf{.023} &
		\textbf{.900} &
		\underline{.867} &
		\underline{.836} &
		\underline{.051} \\
		Ours (UTA)& \textbf{43} &
		\textbf{.915} &
		\textbf{.903} &
		\textbf{.883} &
		\textbf{.037} &
		\textbf{.932} &
		\textbf{.917} &
		\textbf{.905} &
		\textbf{.020} &
		\textbf{.921} &
		\textbf{.905} &
		\textbf{.887} &
		\textbf{.033} &
		\underline{.922} &
		.888 &
		\underline{.865} &
		\underline{.024} &
		\underline{.897} &
		\textbf{.872} &
		\textbf{.843} &
		\textbf{.048} \\ \hline
	\end{tabular}
 	}
	}
\end{table*}
\linespread{1}

\begin{figure*}[t]
	\begin{center}
		%\fbox{\rule{0pt}{2in} \rule{.9\linewidth}{0pt}}
		\includegraphics[width=\textwidth]{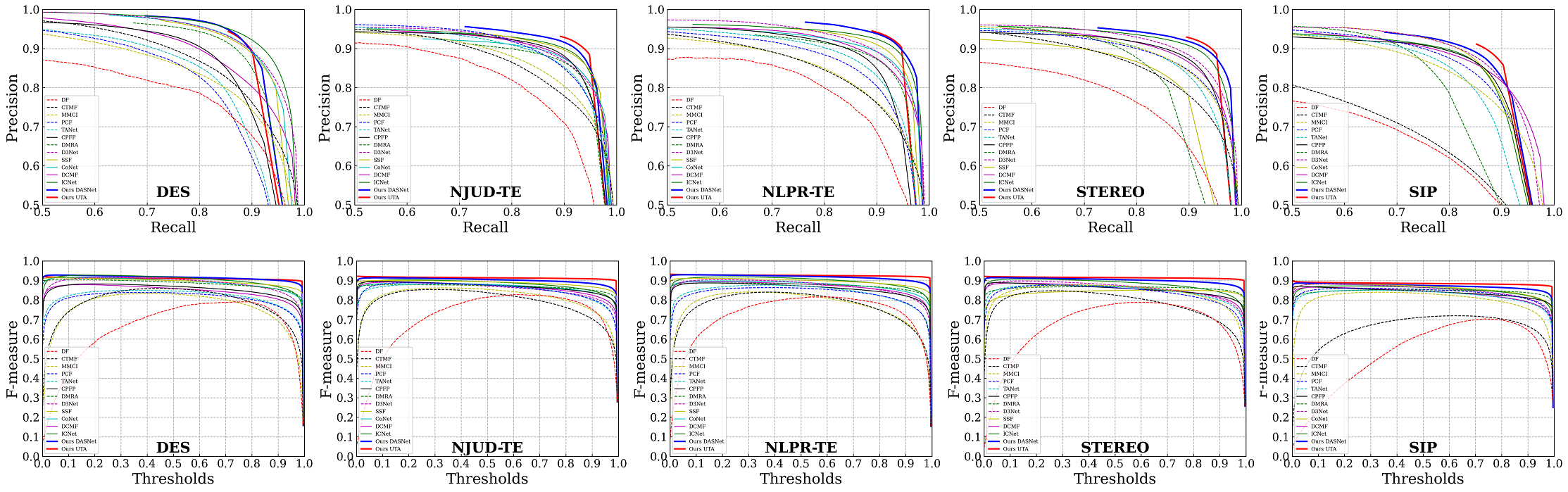}
		\caption{The Precision-Recall curves and F-measure curves of 12 state-of-the-art models and our approaches are listed across five public benchmarks.
		}\label{fig:PR}
	\end{center}
\end{figure*}

\begin{figure*}[t]
	\begin{center}
		%\fbox{\rule{0pt}{2in} \rule{.9\linewidth}{0pt}}
		\includegraphics[width=\linewidth]{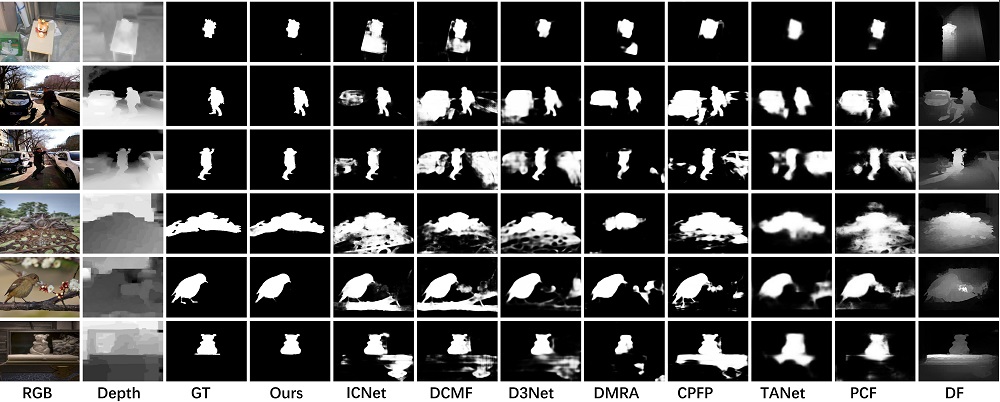}
		\caption{Qualitative comparison of the state-of-the-art RGB-D methods and our approach. Saliency maps produced by our model are clearer and more accurate than others in various challenging scenarios.
		}\label{fig:RGB-D comparison}
	\end{center}
\end{figure*}

\section{EXPERIMENTS}\label{sec:experiment}

\subsection{Datasets and Evaluation Metrics}
\textbf{RGB-D SOD Datasets.} To evaluate the performance of the proposed approach, we conduct experiments on five RGB-D benchmarks~\cite{ju2014depth,peng2014rgbd,niu2012leveraging,cheng2014depth,zhu2017three}, including NJUD~\cite{ju2014depth} with 1,985 images captured by Fuji W3 stereo camera, NLPR~\cite{peng2014rgbd} with 1,000 images captured by Kinect, STEREO~\cite{niu2012leveraging} with 1,000 images collected from the Internet,  DES~\cite{cheng2014depth} with 135 images captured by Kinect, SIP~\cite{fan2019D3Net} with 929 images of human activities captured by Huawei Mate10. Following previous works~\cite{zhao2019contrast,piao2019depth}, we split 1,500 samples from NJUD and 700 samples from the NLPR dataset for training, the rest images in these two datasets and the other three datasets for testing.

\textbf{RGB SOD Datasets.} To verify the effectiveness of our method on RGB datasets, we adopt five RGB benchmarks~\cite{wang2017learning,yang2013saliency,yan2013hierarchical,li2014secrets,li2015visual}, including DUTS~\cite{wang2017learning} with 15,572 images, ECSSD~\cite{yan2013hierarchical} with 1,000 images, DUT-OMRON~\cite{yang2013saliency} with 5,168 images, PASCAL-S~\cite{li2014secrets} with 850 images, HKU-IS~\cite{li2015visual} with 4,447 images. DUTS is currently the largest SOD dataset, following~\cite{wang2017learning}, we split 10,553 images (DUT-TR) from DUTS for training and 5,019 images (DUT-TE) from DUTS for testing, the other four datasets are also used for testing.

\textbf{Evaluation Metrics.} To quantitatively evaluate the performance of our approach and state-of-the-art methods, we adopt 4 commonly used metrics: max F-measure ($F_\beta^{max}$), mean F-measure ($F_\beta^{mean}$), weighted F-measure ($F_\beta^{w}$), mean absolute error ($MAE$), Precision-Recall (PR) curve and F-measure curve on both RGB methods and RGB-D methods.

Following previous works~\cite{oursMM}, We use $F_\beta$ to measure both Precision and Recall. $F_\beta$ is computed based on Precision and Recall pairs as follows:
 \begin{equation} \label{eq:F_beta}
 \begin{split}
 F_\beta = \frac{(1+\beta^2)\times Precision\times Recall}{\beta^2\times Precision + Recall},
 \end{split}
 \end{equation}
 where we set $\beta^2$=0.3 to emphasize more on Precision than Recall, and compute $F_\beta^{max}$, $F_\beta^{mean}$ and $F_\beta^{w}$ using different thresholds as in~\cite{borji2015salient}.

%  We use $S_\alpha$ to measure structure similarity for a more comprehensive evaluation. $S_\alpha$ combines the region-aware ($S_r$) and object-aware ($S_o$) structural similarity as follows:
%   \begin{equation} \label{eq:S_alpha}
%  \begin{split}
%  S_\alpha = \alpha \times S_o + (1-\alpha)\times S_r,
%  \end{split}
%  \end{equation}
%  where we set $\alpha$=0.5 as suggested in~\cite{fan2017structure}.

\subsection{Implementation Details}

We adopt ResNet-50~\cite{he2016deep} pre-trained on ImageNet~\cite{deng2009imagenet} dataset as our backbone. The atrous rate of ASPP follows the prior work~\cite{chen2017deeplab}, which is set as (2, 4, 6). In the training stage, we resize each image to $352\times 352$ and rescale them with the width of $[224,
256, 288, 320, 352]$ for GMS training. Besides, we adopt horizontal flipping, random cropping for data augmentation.  We use the SGD optimizer with the batch size=32 for 48 epochs. Inspired by~\cite{F3Net,chen2020global}, we adopt warm-up and linear decay strategies to adjust the learning rate with the maximum learning rate of 0.005 for ResNet-50 backbone and 0.05 for others. We set the momentum and decay rate as 0.9 and 5e-4 respectively. Our model is lightweight and all experiments are conducted on a single NVIDIA 1080Ti GPU.  The error window size is set as 1 for simplicity.

For the RGB-D SOD, we utilize both RGB images and depth maps from training sets to train our model. During the testing stage, we only need RGB images as inputs to predict saliency maps on RGB-D test sets. For the RGB SOD task, we first estimate depth maps for DUT-TR by pre-trained VNLNet~\cite{yin2019enforcing} directly, which works well in the single image depth estimation task. Then we utilize both DUT-TR and its corresponding predicted depth maps to train our model. During the inference stage, we only need RGB images as inputs to predict saliency maps on RGB test sets.
%The PyTorch implementation will be publicly available.\footnote{Link is masked for blind review policy.}

\subsection{Comparisons with the state-of-the-art}

 As shown in \tabref{table: rgbd benchmark}, we compare our Ubiquitous Target Awareness (UTA) model on 5 widely-used RGB-D SOD benchmarks with 12 state-of-the-art methods, including DF~\cite{qu2017rgbd}, CTMF~\cite{han2017cnns}, MMCI~\cite{chen2019multi}, PCF~\cite{chen2018progressively}, TANet~\cite{chen2019three}, CPFP~\cite{zhao2019contrast}, DMRA~\cite{piao2019depth}, D3Net~\cite{fan2019D3Net}, SSF~\cite{zhang2020select}, CoNet~\cite{Wei2020ECCV}, DCMF~\cite{chen2020rgbd}, ICNet~\cite{li2020icnet}.
 Different from conventional RGB-D SOD methods, it is clear that our method achieves a new performance leader-board without depth images as input, which puts our model in inferior places for comparison.
Especially for the $F_\beta^{max}$, $F_\beta^{mean}$ and $F_\beta^{w}$ metrics, our model outperforms over $3\%$, which means our method has a strong capability to utilize depth information for more precise saliency maps.
In addition, based on our previous work (DASNet)~\cite{oursMM}, the proposed UTANet achieves higher performance on 4 benchmark datasets, while reaches comparable performance on the DES dataset. Beneficial from the design of depth-awareness architecture and lightweight implementation of the SPM, our model performs 43 images per second with a $352 \times 352$ input, reaching a satisfactory trade-off of the speed and accuracy.

%Our proposed approach surpasses 12 state-of-the-art RGB-D salient object detection methods on five public benchmarks by a large margin.

In \figref{fig:RGB-D comparison}, we exhibit results predicted by our model and other approaches. Among these methods, our model performs best on completeness and clarity. The first row exhibit typical scenarios where the salient object is relatively hard to disentangle. Thus with the help of depth awareness, object saliency can be easily deducted.
The second and third rows exploit two complex scenarios where the salient objects are confused with the backgrounds. It can be seen that our model has the potential to detect the most salient object while alleviating ambiguous backgrounds.
For the last three rows, we explore to see what will happen when models facing inferior depth input. It can be seen that simply using depth information cannot detect the holistic object boundary, thus existing models fail to judge the saliency due to their strong dependency on clear depth inputs. In other words, these models tend to give high confidence in the depth boundaries. Unlike these conventional methods, the proposed UTANet shows a clear contextual understanding of these two modalities and is simultaneously aware of low-level cues,~\eg, clear depth boundaries.

\subsection{Analysis of depth quality in RGB-D SOD}

\begin{figure}[t]
	\begin{center}
		%\fbox{\rule{0pt}{2in} \rule{.9\linewidth}{0pt}}
		\includegraphics[width= \linewidth]{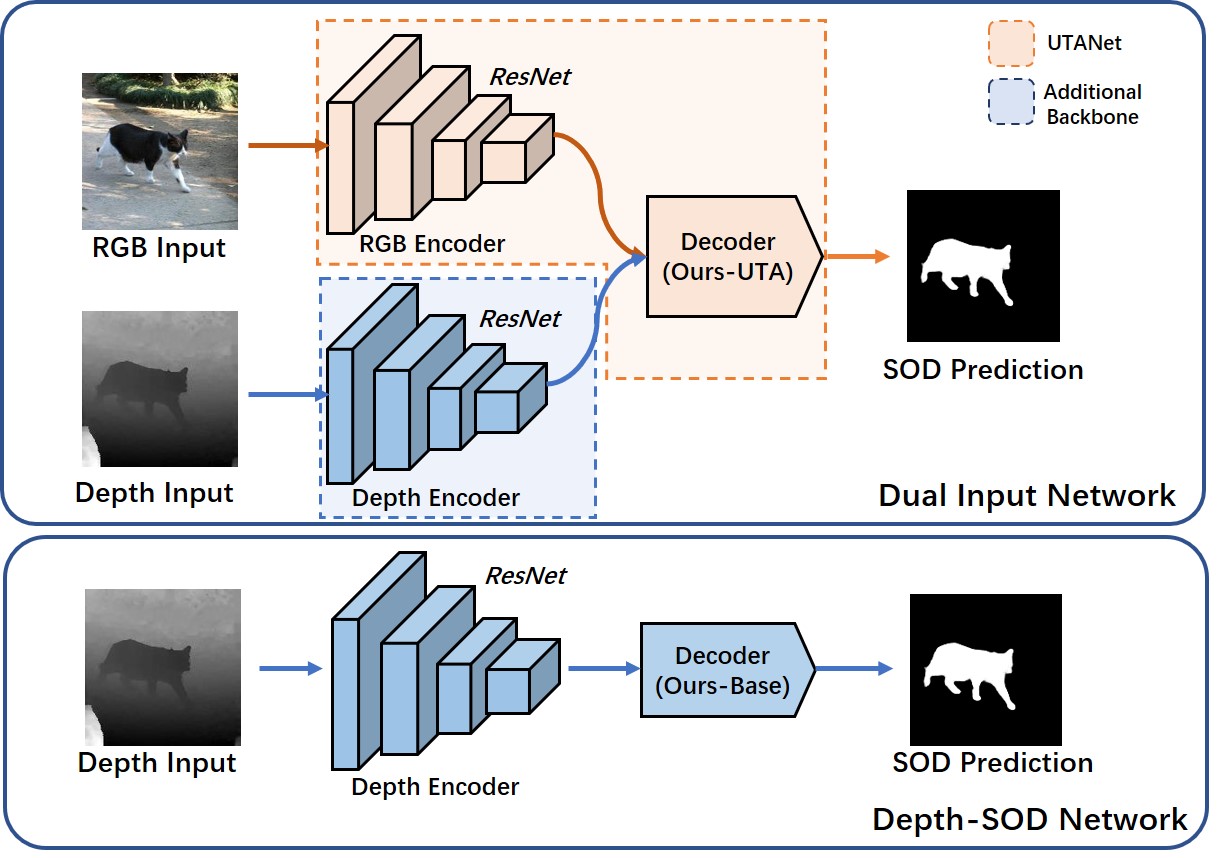}
		\caption{ Network architectures for calculating depth-saliency consistency. Dual input network indicates adding an individual encoder on UTANet to encode depth GT. The Depth-SOD network indicates using only depth information for SOD prediction.
		}\label{fig:qualitynetwork}
	\end{center}
\end{figure}

\begin{table*}[t]
    \centering{
	\caption{Performance comparisons of RGB-D SOD results with different proportions of depth data on 4 benchmarks. Training data are ranked by the quality of depth maps.}
	\label{table:depth_quality}
 	\resizebox{\textwidth}{!}{
 \renewcommand{\arraystretch}{1.3}
	\begin{tabular}{c|c|cccc|cccc|cccc|cccc}
		\hline
        \multirow{2}{*}{Training data} &
		\multicolumn{1}{c|}{\multirow{2}{*}{METHODS}} &
        \multicolumn{4}{c|}{NJUD-TE} &
        \multicolumn{4}{c|}{NLPR-TE} &
		\multicolumn{4}{c|}{STEREO} &
		\multicolumn{4}{c}{SIP} \\
    & & $F_\beta^{max}$ & $F_\beta^{mean}$  & $MAE$ & $F_\beta^{w}$& $F_\beta^{max}$ & $F_\beta^{mean}$  & $MAE$ & $F_\beta^{w}$ & $F_\beta^{max}$ & $F_\beta^{mean}$  & $MAE$ & $F_\beta^{w}$& $F_\beta^{max}$ & $F_\beta^{mean}$  & $MAE$ & $F_\beta^{w}$
\\
         \hline
		\multirow{2}{*}{Top 30\%}& Ours (Depth-Aware)& .856 &.830 &.064 &.802  &.919&.902 &.023 &.890  & .879 &.854 &.048 &.831  &.833&.796 &.078 &.754 \\
		&Ours (Depth GT)&.869&.844&.058&.819 &.932&.920&.018&.909  &.891&.864&.047&.841 &.877&.846&.062&.809  \\
        \hline
		\multirow{2}{*}{Top 50\%}& Ours (Depth-Aware)& .879 &.858 &.053 &.837  &.826&.911 &.021 &.898   & .890 &.867 &.044 &.846  & .855 &.825 &.067 &.786\\
		&Ours (Depth GT)&.911&.892&.041&.875  &.935&.919&.019&.909  &.905&.883&.040&.863 &.892&.863&.053&.833 \\
        \hline
		\multirow{2}{*}{Top 70\%}& Ours (Depth-Aware)& .898 &.878 &.046 &.857   &.934&.919 &.019 &.907  & .912 &.894 &.036 &.876  &.900&.874 &.050 &.842 \\
		&Ours (Depth GT) &.911&.892&.041&.875  &.935&.919&.019&.909  &.921&.900&.035&.882  &.905&.881&.045&.854 \\
        \hline
		\multirow{2}{*}{Top 100\%}& Ours (Depth-Aware)& .915 &.903 &.037 &.883   &.932&.917 &.020 &.905   & .921 &.905 &.033 &.887   &.897&.872 &.048 &.843\\
		&Ours (Depth GT)&.927&.913&.034&.898  &.933&.918&.019&.907  &.920&.899&.034 &.883  &.903&.878&.048&.849 \\
\hline
	\end{tabular}
 	}
	}
\end{table*}

With the aforementioned phenomenon, we summarize that the unstable depth quality would provide an inferior understanding of salient objects. To verify the challenge in~\figref{fig:depth_dis}, we make an experimental study to explore the relationship between depth quality and the SOD learning process.

\textbf{Evaluation setting.} The clear depth groundtruth of these low-quality images cannot be directly obtained. Inspired by existing work~\cite{fan2019D3Net} which reduces the low-quality depth maps for learning, we define a ``depth-saliency consistency" as the consistency of the transformed depth maps with the salient object detections results. In other words, if the depth map can directly predict the SOD mask, it provides useful cues for the SOD learning process. We then adopt the depth-SOD encoding network in~\figref{fig:qualitynetwork} with backbone encoder and a single-branch decoder from UTANet (denoted as \textit{Ours-Base}). In addition, the predictions of salient objects are probability distributions. Thus instead of comparing the training prediction with 0-1 groundtruth masks, we measure them with the prediction of the dual input network in~\figref{fig:qualitynetwork}.

To measure the ``depth-saliency consistency" scores, we use the MAE metric for evaluations.  In this manner, we re-partition the dataset with similarity ranking scores and extract the top 30\% to 100\% of the training data.  We compare two different types of network: the proposed UTANet (namely depth-aware) with only RGB input for inference, and the dual input network by adding an auxiliary depth encoder on UTANet.
We conduct comparisons in~\tabref{table:depth_quality} on four widely-used RGB-D benchmarks. From these results, several interesting findings are concluded:
1) High-quality depth data can facilitate the learning process of salient object detection. In the setting of top 30\% and 50\% data, it can be found that by adding depth GT inputs with an additional encoder, the performance is improved consistently on almost all evaluation metrics.
2) With the increase of inferior depth data, the network with depth GT performs similar results with the proposed UTANet. By observing the F-max value of SIP dataset, with the accurate depth cues for SOD, the depth-awareness network performs much lower results than that with depth GT,~\eg, 0.833 and 0.877 in $F_\beta^{max}$. However, when utilizing all training data, the performance of both settings is increased but the gap between these methods become close,~\eg, 0.897 and 0.903 in $F_\beta^{max}$.

In summary, it is concluded that the proposed UTANet handles the different quality of training depth well and is independent of depth quality. While the dual input network performs well with high-quality depth data but leads to inferior results with low-quality depth data.

\begin{table*}[t]
\caption{Ablation study for different components. CAF denotes the proposed channel aware fusion module. SPM denotes the spatial perceptive module, DAC denotes the depth awareness constraint. GMS denotes the proposed gated multi-scale predictor. BCE, IoU, DEC are different loss functions mentioned above.  MLS represents multi-level supervision. }
\label{table:ablation}
\begin{center}
\renewcommand\tabcolsep{4.0pt}
\begin{tabular}{ccccc|ccc|ccc|ccc|ccc|c}
\hline
\multirow{2}{*}{DAC} &
  \multirow{2}{*}{SPM} &
  \multirow{2}{*}{CAF} &
  \multirow{2}{*}{GMS} &
  \multirow{2}{*}{MLS} &
  \multicolumn{3}{c|}{NJUD-TE} &
  \multicolumn{3}{c|}{NLPR-TE} &
  \multicolumn{3}{c|}{STERE} &
  \multicolumn{3}{c|}{SIP} &
  \multirow{2}{*}{FPS} \\
  &   &   &   &   & $F^{mean}_\beta$ & $F^{w}_\beta$ & $MAE$ & $F^{mean}_\beta$ & $F^{w}_\beta$ & $MAE$ & $F^{mean}_\beta$ & $F^{w}_\beta$ & $MAE$ & $F^{mean}_\beta$ & $F^{w}_\beta$ & $MAE$  &    \\ \hline
  &   &   &   &   & .855 & .830 & .052 & .856 & .850 & .029 & .856 & .833 & .047 & .826 & .788 & .062 & 80 \\
 \checkmark &   &   &   &   & .863 & .837 & .051 & .868 & .860 & .027 & .867 & .843 & .044 & .836 & .799 & .059 & 65 \\
 \checkmark & \checkmark &   &   &   & .882 & .862 & .042 & .884 & .878 & .024 & .887 & .868 & .037 & .857 & .823 & .052 & 59 \\
  &   & \checkmark &   &   & .880 & .861 & .044 & .884 & .876 & .025 & .884 & .865 & .038 & .856 & .826 & .051 & 58 \\
\checkmark  &   & \checkmark &   &   & .889 & .870 & .042 & .892 & .880 & .024 & .888 & .867 & .038 & .864 & .835 & .049 & 48 \\
\checkmark & \checkmark & \checkmark &   &   & .891 & .871 & .040 & .908 & .894 & .021 & .893 & .874 & .036 & .868 & .834 & .049 &  44 \\
\checkmark & \checkmark & \checkmark & \checkmark &   & .896 & .875 & .039 & .917 & .904 & \textbf{.020} & .899 & .878 & .035 & \textbf{.874} & \textbf{.844} & \textbf{.048} & 43 \\
\checkmark & \checkmark & \checkmark & \checkmark & \checkmark & \textbf{.903} & \textbf{.883} & \textbf{.037} & \textbf{.917} & \textbf{.905} & \textbf{.020} & \textbf{.905} & \textbf{.887} & \textbf{.033} & .872 & .843 & \textbf{.048} & 43 \\ \hline
\end{tabular}
\end{center}
\end{table*}
\linespread{1}

\begin{figure}[t]
	\begin{center}
		%\fbox{\rule{0pt}{2in} \rule{.9\linewidth}{0pt}}
		\includegraphics[width= \linewidth]{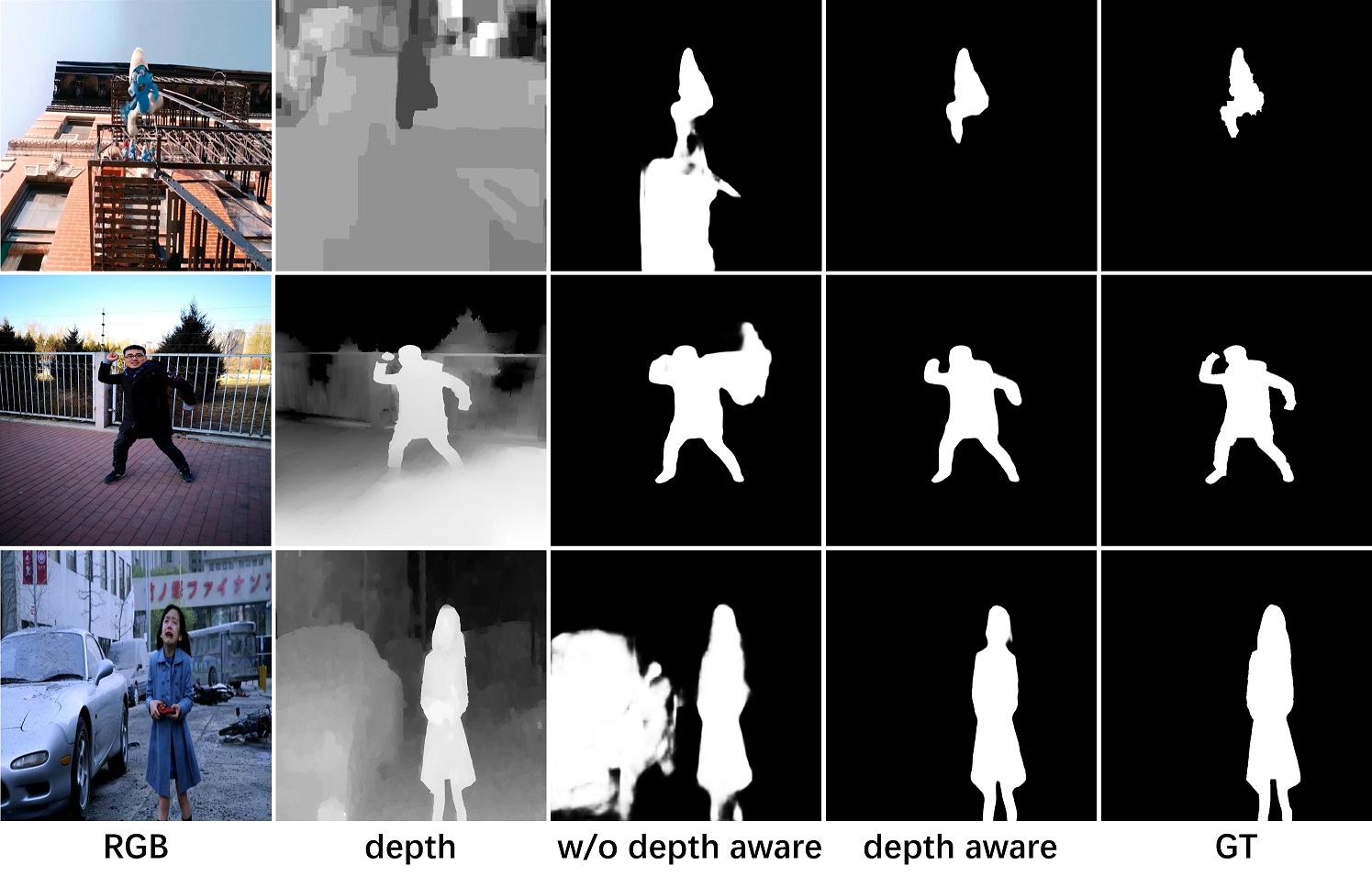}
		\caption{Qualitative comparisons on RGB-D datasets. The third column without depth awareness is hard to distinguish complex scenarios with similar foreground and background, while our model in the fourth column shows better performance.
		}\label{fig:analysis}
	\end{center}
\end{figure}

% ablation of SPM details
\linespread{1.3}
\begin{table}[t]
    \caption{Ablation study for cross-modal fusion strategies. All-CAF: replacing SPM module with CAF module for cross-modalities. }
\label{table: SPM ablation}
\begin{center}
\renewcommand\tabcolsep{3.0pt}
\begin{tabular}{c|cc|cc|cc}
\hline
\multirow{2}{*}{} & \multicolumn{2}{c|}{NJUD-TE} & \multicolumn{2}{c|}{NLPR-TE} & \multicolumn{2}{c}{STERE}  \\
         & $F^{mean}_\beta$ & $MAE$ & $F^{mean}_\beta$ & $MAE$ & $F^{mean}_\beta$ & $MAE$ \\ \hline
UTA (full model)     & .903 & .037 & .917 & .020 & .905 & .033  \\
w/o SPM  & .890 & .041 & .909 & .020 & .893 & .033  \\
All-CAF     & .894 & .041 & .908 & .022 & .899 & .036  \\
w/o Edge & .897 & .038 & .912 & .021 & .898 & .034  \\\hline
\end{tabular}
\end{center}
\end{table}
\linespread{1}

\subsection{Performance Analysis}

To investigate the effectiveness of each key component in our proposed model, we first conduct a thorough ablation study and then measure the computation complexity. After that, a comparison on RGB SOD benchmarks is exhibited to verify the extensibility of our model.

\textbf{Ablation Studies.} To evaluate the effectiveness of our feature fusion module, we reconstruct our model with different ablation factors in~\tabref{table:ablation}. We first replace the cross-level fusion module with multiplication operations as our baseline in the first row, which only includes the IoU Loss and BCE loss. Note that this baseline model outperforms several state-of-the-art models and our proposed model can steadily improve the performance based on this high baseline.
As an important constraint to the SOD feature, using depth awareness constraints (denoted as DAC) in a multi-task learning trend can boost the performance. Although the DAC module includes the auxiliary depth branch and depth error correction (DEC), the relation between multiple modalities is unexplored.
To solve this, we add the cross-modal fusion strategies SPM, the performance improves notably,~\eg, over $2\%$ in $F_\beta^{mean}$ and $F_\beta^{w}$ measurements.

\textbf{Channel-aware Cross-level Interaction.} Our channel-aware cross-level interaction is denoted as CAF in~\tabref{table:ablation}, adopting channel-aware attention is proved to be necessary for fusing and selecting channels from multiple levels, which can boost over $3\%$ performance in $F_\beta^{mean}$ and $F_\beta^{w}$ compared to the high baseline model.
At last, we add multi-level supervision to refine our results. As shown in \tabref{table:ablation}, all components contribute to the performance improvement, which demonstrates the necessity of each component of our proposed model to obtain the best saliency detection results.

\textbf{Depth-awareness Constraints.} Qualitative results can be found in~\figref{fig:analysis}. The SOD results with and without depth awareness constraints are presented in the third and fourth columns. It is interesting to find that simply using the RGB image, the salient objects are hard to localize,~\eg, the person in the first image. However, these objects are salient in the depth map and provide clear localization cues. Therefore, using depth-awareness constraints is beneficial for the salient localization in the fourth column.

\linespread{1.3}
\begin{table}[]
\caption{Ablations of the proposed gated multi-scale predictor. The red arrows {\color{red}$(\uparrow)$} and blue arrows {\color{blue}$(\downarrow)$} denote the relative improvements respectively. }
\label{table: GMS generalization}
\begin{center}
% 	\resizebox{\columnwidth}{!}{
\renewcommand\tabcolsep{1.0pt}
\begin{tabular}{c|cc|cc|cc}
\hline
\multirow{2}{*}{Methods} & \multicolumn{2}{c|}{NJUD-TE} & \multicolumn{2}{c|}{NLPR-TE} & \multicolumn{2}{c}{STERE} \\
      & $F^{mean}_\beta$ & $MAE$ & $F^{mean}_\beta$ & $MAE$ & $F^{mean}_\beta$ & $MAE$ \\ \hline
baseline & .855 & .052 & .856 & .029 & .856 & .047  \\
+GMS  & .874{\color{red}$(\uparrow1.9)$} & .046 & .886{\color{red}$(\uparrow3.0)$} & .025 & .876{\color{red}$(\uparrow2.0)$} & .040  \\
+PPM  & .862{\color{red}$(\uparrow0.7)$} & .051 & .868{\color{red}$(\uparrow1.2)$} & .027 & .868{\color{red}$(\uparrow1.2)$} & .043  \\
+MSI  & .857{\color{red}$(\uparrow0.2)$} & .054 & .848{\color{blue}$(\downarrow0.8)$} & .030 & .857{\color{red}$(\uparrow0.1)$} & .047  \\ \hline
Ours w/o GMS  & .898 & .040 & .909 & .020 & .897 & .036  \\
+GMS  & .903{\color{red}$(\uparrow0.5)$} & .037 & .917{\color{red}$(\uparrow0.8)$} & .020 & .905{\color{red}$(\uparrow0.8)$} & .033 \\
+PPM  & .895{\color{blue}$(\downarrow0.3)$} & .040 & .900{\color{blue}$(\downarrow0.9)$} & .023 & .898{\color{red}$(\uparrow0.1)$} & .035  \\
+MSI  & .889{\color{blue}$(\downarrow0.9)$} & .042 & .899{\color{blue}$(\downarrow1.0)$} & .021 & .893{\color{blue}$(\downarrow0.4)$} & .037 \\ \hline
UCNet \cite{zhang2020uc} & .885 & .044 & .884 & .028 & .874 & .041 \\
+GMS  & .887{\color{red}$(\uparrow0.2)$}  & .045 & .888{\color{red}$(\uparrow0.4)$} & .027 & .885{\color{red}$(\uparrow1.1)$} & .038 \\
+PPM  & .877{\color{blue}$(\downarrow0.8)$} & .047 & .886{\color{red}$(\uparrow0.2)$} & .027 & .869{\color{blue}$(\downarrow0.5)$} & .044 \\
+MSI  & .871{\color{blue}$(\downarrow1.4)$}  & .046 & .861{\color{blue}$(\downarrow2.3)$} & .031 & .865{\color{blue}$(\downarrow0.9)$} & .044 \\ \hline
\end{tabular}
% }
\end{center}
\end{table}
\linespread{1}

\linespread{1.2}
\begin{table*}[t]
    \centering{
	\caption{Performance comparison with 10 state-of-the-art RGB SOD methods on five benchmarks. Smaller $MAE$, larger $F_\beta^{max}$,$F_\beta^{mean}$ and $F_\beta^{w}$ correspond to better performance. The best results are highlighted in bold.}
	\label{table:RGB benchmark}
% 	\resizebox{\textwidth}{!}{
	 \renewcommand\tabcolsep{1.0pt}
		\begin{tabular}{l|c|cccc|cccc|cccc|cccc|cccc}
			\hline
			\multicolumn{1}{c|}{\multirow{2}{*}{METHODS}} &
			\multirow{2}{*}{FPS} &
			\multicolumn{4}{c|}{ECSSD} &
			\multicolumn{4}{c|}{DUT-TE} &
			\multicolumn{4}{c|}{DUT-OMRON} &
			\multicolumn{4}{c|}{HKU-IS} &
			\multicolumn{4}{c}{PASCAL-S} \\ %\cline{2-21}
			\multicolumn{1}{c|}{} & &
			$F_\beta^{max}$ &
			$F_\beta^{mean}$ &
			$F_\beta^{w}$ &
			$MAE$ &
			$F_\beta^{max}$ &
			$F_\beta^{mean}$ &
			$F_\beta^{w}$ &
			$MAE$ &
			$F_\beta^{max}$ &
			$F_\beta^{mean}$ &
			$F_\beta^{w}$ &
			$MAE$ &
			$F_\beta^{max}$ &
			$F_\beta^{mean}$ &
			$F_\beta^{w}$ &
			$MAE$ &
			$F_\beta^{max}$ &
			$F_\beta^{mean}$ &
			$F_\beta^{w}$ &
			$MAE$ \\ \hline
			BMPM~\cite{zhang2018bi} & 28 &
			.929 &
			.894 &
			.871 &
			.045 &
			.851 &
			.762 &
			.761 &
			.049 &
			.774 &
			.698 &
			.681 &
			.064 &
			.921 &
			.875 &
			.860 &
			.039 &
			.862 &
			.803 &
			.785 &
			.073 \\
			PAGR~\cite{zhang2018progressive} & - &
			.927 &
			.894 &
			.834 &
			.061 &
			.854 &
			.784 &
			.724 &
			.056 &
			.771 &
			.711 &
			.622 &
			.071 &
			.918 &
			.886 &
			.823 &
			.048 &
			.854 &
			.803 &
			.738 &
			.094 \\
			R3Net~\cite{deng2018r3net} & 22 &
			.934 &
			.914 &
			.902 &
			.040 &
			- &
			- &
			- &
			- &
			.795 &
			.747 &
			.728 &
			.062 &
			.915 &
			.893 &
			.877 &
			.036 &
			.842 &
			.800 &
			.760 &
			.095 \\
			PiCA-R~\cite{liu2018picanet} & 5 &
			.935 &
			.901 &
			.867 &
			.046 &
			.860 &
			.759 &
			.755 &
			.051 &
			.803 &
			.717 &
			.695 &
			.065 &
			.919 &
			.880 &
			.842 &
			.043 &
			.867 &
			.800 &
			.782 &
			.077 \\
			BANet~\cite{su2019selectivity} & - &
			.939 &
			.917 &
			.908 &
			.041 &
			.872 &
			.829 &
			.811 &
			.040 &
			.782 &
			.750 &
			.736 &
			.061 &
			.923 &
			.893 &
			.887 &
			.037 &
			.847 &
			.839 &
			.817 &
			.079 \\
			PoolNet~\cite{liu2019simple} & 32 &
			.944 &
			.915 &
			.896 &
			.039 &
			.880 &
			.809 &
			.807 &
			.040 &
			.808 &
			.747 &
			.729 &
			.055 &
			.933 &
			.899 &
			.883 &
			.032 &
			.869 &
			.822 &
			.800 &
			.074 \\
			BASNet~\cite{qin2019basnet} & 25 &
			.943 &
			.880 &
			.904 &
			.037 &
			.859 &
			.791 &
			.803 &
			.048 &
			.805 &
			.756 &
			.751 &
			.056 &
			.928 &
			.895 &
			.890 &
			.032 &
			.857 &
			.775 &
			.800 &
			.078 \\
			CPD-R~\cite{wu2019cascaded} & \textbf{66} &
			.939 &
			.917 &
			.898 &
			.037 &
			.865 &
			.805 &
			.795 &
			.043 &
			.797 &
			.747 &
			.719 &
			.056 &
			.925 &
			.891 &
			.876 &
			.034 &
			.864 &
			.824 &
			.803 &
			.072 \\
			F3Net~\cite{F3Net} & - &
			.945 &
			.925 &
			.912 &
			.033 &
			\textbf{.890} &
			.840 &
			.835 &
			\textbf{.035} &
			.813 &
			.766 &
			.746 &
			.053 &
			.937 &
			.910 &
			.900 &
			.028 &
			.880 &
			.840 &
			.821 &
			\textbf{.064} \\
			GCPANet~\cite{chen2020global} & - &
			.948 &
			.919 &
			 .903&
			.035 &
			.888 &
			.817 &
			.821 &
			.040 &
			.812 &
			.748 &
			.734 &
			.056 &
			.938 &
			.898 &
			.889 &
			.031 &
			.876 &
			.836 &
			.816 &
			\textbf{.064} \\ \hline
			
			%Ours DASNet \cite{}& 33 &.950 &.932 &.918 &.032 &.896 &.853 &.843 &.034 &.827 &.783 &.767 &.050 &.942 &.917 &.907 &.027 &.885 &.849 &.828 &.064 \\
			Ours (UTA) & \underline{43} &
			\textbf{.952} &
			\textbf{.937} &
			\textbf{.924} &
			\textbf{.030} &
			\textbf{.890} &
			\textbf{.862} &
			\textbf{.849} &
			\textbf{.035} &
			\textbf{.816} &
			\textbf{.781} &
			\textbf{.766} &
			\textbf{.048} &
			\textbf{.939} &
			\textbf{.921} &
			\textbf{.910} &
			\textbf{.026} &
			\textbf{.882} &
			\textbf{.852} &
			\textbf{.828} &
			\textbf{.064} \\ \hline
	\end{tabular}
% 	}
	}
\end{table*}

\textbf{Spatial-aware Cross-modal Interaction.}
Are spatial cues necessary for cross-modality interactions? To verify the effectiveness of our proposed spatial perceptive module, we conduct detailed ablations in~\tabref{table: SPM ablation}.
The first setting in the second row is to replace all SPM modules with the feature concatenation operations (denoted as w/o SPM). It can be found that the performance drop by about $1\%$ when missing the positional information. The second setting denoted as (ALL-CAF) utilizes the efficient CAF module to replace the original SPM module, which leads to inferior performance.
Last but not least, we remove the edge attention module (w/o Edge), the module achieves better performance than the first two settings but lower performance than our full model. This verifies that spatial information is thus necessary for cross-modality fusion.

\textbf{Gated Multi-scale Predictor.}
% verify improvement in our net
To compare with other typical multi-scale strategies,~\ie, pyramid pooling module (PPM), multi-scale inference (MSI) as mentioned in~\secref{sec:gms}, we conduct a thorough experiment in~\tabref{table: GMS generalization}. It is verified that both our GMS and PPM module can notably improve the quality of saliency detection results,~\ie, $1\%$ of PPM and over $2\%$ of our GMS module. In the second block, we exhibit the results of our full model without the GMS module. Based on this high-performance module, PPM and MSI fail to refine the segmentation results.
As a comparison, our proposed GMS can still boost the $F_\beta^{mean}$ by about $0.8\%$ on this high base.

% verify improvement and robustness in other net
Our gated multi-scale predictor is easy to plug into the existing salient object detection networks. To verify the generalization ability of GMS, we plug our gated multi-scale predictor, pyramid pooling module, and multi-scale inference strategy into other networks respectively. The training and testing protocol follows the official code in~\cite{zhang2020uc}.
As shown~\tabref{table: GMS generalization} shows, our gated multi-scale predictor could achieve an explicit improvement in all experiments and achieve higher performance than the other two multi-scale aggregation strategies. This extended experiment demonstrates the generalization ability of our gated multi-scale predictor.

\linespread{1}
\begin{figure}[t]
	\begin{center}
		%\fbox{\rule{0pt}{2in} \rule{.9\linewidth}{0pt}}
		\includegraphics[width=1\linewidth]{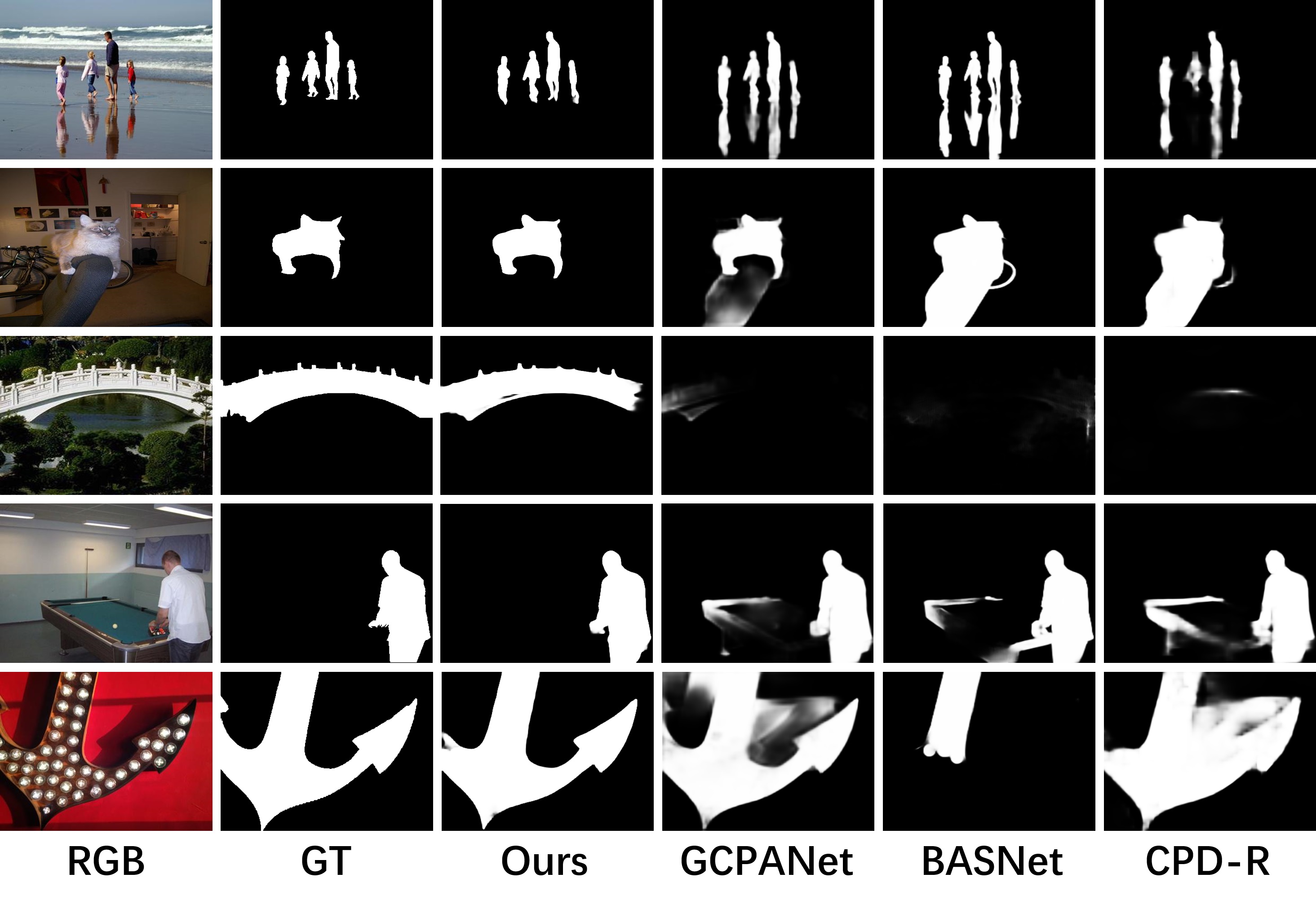}
		\caption{Qualitative comparison of the state-of-the-art RGB methods and our approach. Obviously, saliency maps produced by our model are clearer and more accurate than others in various challenging scenarios.
		}\label{fig:RGB comparison}
	\end{center}
\end{figure}

\textbf{Extensions on RGB SOD Benchmark.}
As a depth-awareness model, our testing phase does rely on the captured depth groundtruth. Thus we utilize a coarse estimated depth map for depth supervision and verify the flexibility and extensibility on RGB SOD tasks.
As shown in \tabref{table:RGB benchmark}, we compare our proposed UTA with 10 state-of-the-art methods,~\ie, BMPM~\cite{zhang2018bi}, PAGR~\cite{zhang2018progressive}, R3Net~\cite{deng2018r3net}, PiCANet~\cite{liu2018picanet}, PoolNet~\cite{liu2019simple}, BANet~\cite{su2019selectivity}, CPDNet~\cite{wu2019cascaded}, BASNet~\cite{qin2019basnet}, F3Net~\cite{F3Net}, and GCPANet~\cite{chen2020global}.
As shown in~\tabref{table:RGB benchmark}, we can see our proposed UTA still outperforms other methods and ranks first on all datasets and almost all metrics. However, this performance is achieved with only estimated depth maps as training priors. we believe that with the captured real data, the final performance would be improved steadily, which is validated on the RGB-D benchmarks.
As shown in \figref{fig:RGB comparison}, comparing with the visual results of different methods, our approach shows a clear superiority in both completeness and clarity.
It is can be found that our model precisely captures the salient object and omits the background confusions.

\linespread{1.3}
\begin{table}[t]
\renewcommand\tabcolsep{2.0pt}
	\caption{Complexity comparison with RGB models and RGB-D models. Models ranking the first and second place are viewed in bold and underlined.}
	\label{table:params}
	\begin{tabular}{cccccc}
		\hline
		& Methods & Platform & Params(M)      & MAdds(G)      & FPS \\ \hline
		\multirow{2}{*}{RGB\&RGB-D}
		& Ours UTA & pytorch  & \textbf{31.83} & {\underline{10.27}}   & {\underline{43}} \\
		& Ours DASNet \cite{oursMM}    & pytorch  & 36.68 & 11.57   & 33 \\
		\hline
		\multirow{4}{*}{RGB-D} & CPFP~\cite{zhao2019contrast}    & caffe    & 72.94          & 21.25     & 7      \\
		& DMRA~\cite{piao2019depth}    & pytorch  & 59.66          & 113.09    & 10      \\
		& CoNet~\cite{Wei2020ECCV}    & pytorch  & 43.86          & 34.39    & 34      \\
		& SSF~\cite{zhang2020select}    & pytorch  &     {\underline{32.93}}      & 46.53 & 32      \\ \hline
		\multirow{4}{*}{RGB}  & GCPANet~\cite{chen2020global} & pytorch  & 67.06          & 26.61     & -    \\
		& BASNet~\cite{qin2019basnet}  & pytorch  & 87.06          & 97.51   & 25        \\
		& CPD-R~\cite{wu2019cascaded}   & pytorch  & 47.85    & \textbf{7.19} & \textbf{66}   \\
		& BANet~\cite{su2019selectivity}   & caffe    & 55.90          & 35.83       & -     \\ \hline
	\end{tabular}
\end{table}
\linespread{1}

\textbf{Computational Efficiency.} \tabref{table:params} shows the parameters and computational cost measured by multiply-adds (MAdds) of our proposed model and other open-sourced RGB models and RGB-D models. Our model could achieve obvious higher performance in a lightweight fashion. Notably, SSF~\cite{zhang2020select} and CPD-R~\cite{wu2019cascaded} discard features of two shallower layers to improve the computation efficiency, but sacrifice the accuracy and clarity of results. For fair comparisons, we obtain the deployment codes released by authors and evaluate them with the same configuration.

\section{CONCLUSIONS}\label{sec:conclusion}
In this paper, we revisit the problem of depth in the field of salient object detection and propose a novel depth-awareness architecture to solve this important problem. The proposed depth-awareness setting takes depth as supervision in the network learning stage, to regularize the features to be aware of depth information. Base on this architecture, we propose Ubiquitous Target Awareness (UTA) network to solve the three main challenges in RGB-D SOD tasks,~\ie, depth regularization and error correction, low-level cues awareness in cross-modal fusion, and multi-scale perception. The depth-awareness module first provides depth localization cues to find the saliency consensus of depth and SOD branch and generates the depth-error weights to mine salient ambiguous regions. Second, the predicted depth stream is regularized by a spatial perceptive module, being aware of boundary cues simultaneously. Besides, we propose a generalized multi-scale gating module for salient object detection, which is implementation-friendly and can be plugged into multiple existing SOD frameworks.
These three modules work collaboratively to promote the learning process of salient object detection.
Experimental evidence reveals that with only RGB inputs, the proposed network not only surpasses the state-of-the-art RGB-D methods by a large margin but well demonstrates its effectiveness in RGB application scenarios.

% or


\section*{Acknowledgment}
This work was supported by grants from National Natural Science Foundation of China (No.61922006) and Baidu academic collaboration program.

\ifCLASSOPTIONcaptionsoff
  \newpage
\fi

%\bibliographystyle{IEEEtran}
%\bibliography{IEEEabrv,../IQA}

\bibliographystyle{IEEEtran}
\bibliography{DESA}

\balance
\end{document}